\documentclass[10pt,table]{article} 
\usepackage[utf8]{inputenc} 
\DeclareUnicodeCharacter{00D7}{\ensuremath{\times}}
\usepackage[accepted]{tmlr}

\usepackage{amsmath,amsfonts,bm}

\def\eqref#1{equation~\ref{#1}}

\def\1{\bm{1}}

\def\vtheta{{\bm{\theta}}}

\DeclareMathAlphabet{\mathsfit}{\encodingdefault}{\sfdefault}{m}{sl}
\SetMathAlphabet{\mathsfit}{bold}{\encodingdefault}{\sfdefault}{bx}{n}

\newcommand{\R}{\mathbb{R}}

\usepackage{hyperref}
\usepackage{url}
\usepackage{acronym}

\acrodef{DNN}{Deep Neural Network}
\acrodefplural{DNN}[DNNs]{Deep Neural Networks}

\acrodef{BNN}{Binarized Neural Network}
\acrodefplural{BNN}[BNNs]{Binarized Neural Networks}

\acrodef{MAC}{Multiply-Accumulate}
\acrodefplural{MAC}[MACs]{Multiply-Accumulates}

\acrodef{DSP}{Digital Signal Processor}
\acrodefplural{DSP}[DSPs]{Digital Signal Processors}

\acrodef{BRAM}{Block Random Access Memory}
\acrodefplural{BRAM}[BRAMs]{Block Random Access Memories}

\acrodef{ROM}{Read-Only Memory}
\acrodefplural{ROM}[ROMs]{Read-Only Memories}

\acrodef{HBM}{High-Bandwidth Memory}

\acrodef{LUT}{Lookup Table}
\acrodefplural{LUT}[LUTs]{Lookup Tables}

\acrodef{MLP}{Multilayer Perceptron}
\acrodefplural{MLP}[MLPs]{Multilayer Perceptrons}

\acrodef{FPGA}{Field-Programmable Gate Array}
\acrodefplural{FPGA}[FPGAs]{Field-Programmable Gate Arrays}

\acrodef{GPU}{Graphics Processing Unit}
\acrodefplural{GPU}[GPUs]{Graphics Processing Units}

\acrodef{CPU}{Central Processing Unit}
\acrodefplural{CPU}[CPUs]{Central Processing Units}

\acrodef{CNN}{Convolutional Neural Network}
\acrodefplural{CNN}[CNNs]{Convolutional Neural Networks}

\acrodef{XNOR}{exclusive NOR}
\acrodefplural{XNOR}[XNORs]{exclusive NORs}

\acrodef{TCL}{Tool Command Language}
\acrodefplural{TCL}[TCLs]{Tool Command Languages}

\acrodef{HDL}{Hardware Description Language}
\acrodefplural{HDL}[HDLs]{Hardware Description Languages}

\acrodef{RTL}{Register Transfer Level}
\acrodefplural{RTL}[RTLs]{Register Transfer Levels}

\acrodef{DUT}{Device Under Test}
\acrodefplural{DUT}[DUTs]{Devices Under Test}

\acrodef{DRC}{Design Rule Check}
\acrodefplural{DRC}[DRCs]{Design Rule Checks}

\acrodef{NAS}{Neural Architecture Search}

\acrodef{II}{Initiation Interval}

\acrodef{FSM}{Finite-State Machine}
\acrodefplural{FSM}[FSMs]{Finite-State Machines}

\acrodef{FF}{Flip-Flop}
\acrodefplural{FF}[FFs]{Flip-Flops}

\acrodef{SAIF}{Switching Activity Interchange Format}

\acrodef{SSM}{State Space Model}
\acrodefplural{SSM}[SSMs]{State Space Models}

\acrodef{FFN}{Feedforward Network}
\acrodefplural{FFN}[FFNs]{Feedforward Networks}

\acrodef{NN}{Neural Network}
\acrodefplural{NN}[NNs]{Neural Networks}

\acrodef{DWN}{Differentiable Weightless Neural Network}
\acrodefplural{DWN}[DWNs]{Differentiable Weightless Neural Networks}

\acrodef{ASIC}{Application-Specific Integrated Circuit}
\acrodefplural{ASIC}[ASICs]{Application-Specific Integrated Circuits}

\acrodef{CUDA}{Compute Unified Device Architecture}

\acrodef{NVML}{NVIDIA Management Library}

\acrodef{DVFS}{Dynamic Voltage and Frequency Scaling}

\acrodef{WiSARD}{Wilkie, Stonham and Aleksander's Recognition Device}

\acrodef{LLM}{Large Language Model}
\acrodefplural{LLM}[LLMs]{Large Language Models}

\usepackage{tikz}
\tikzset{scriptsize/.style={font=\normalsize}}
\usepackage{graphicx}
\usepackage{subcaption}
\usepackage{adjustbox}
\usepackage{colortbl}
\usetikzlibrary{arrows.meta} 
\usepackage{booktabs}
\usepackage{cleveref}
\crefname{equation}{composition}{compositions}
\Crefname{equation}{Composition}{Compositions}
\usepackage{multirow}
\usepackage{threeparttable} 
\usepackage{placeins} 

\title{BitLogic: A Framework for Gradient-Based LUT-Native Neural Networks}

\author{
    \name Simon Bührer \email sbuehrer@ethz.ch \\
    \addr ETH Zurich \\
    Zurich, Switzerland
    \AND
    \name Andreas Plesner \email aplesner@ethz.ch \\
    \addr ETH Zurich \\
    Zurich, Switzerland
    \AND
    \name Till Aczel \email taczel@ethz.ch \\
    \addr ETH Zurich \\
    Zurich, Switzerland
    \AND
    \name Roger Wattenhofer \email wattenhofer@ethz.ch \\
    \addr ETH Zurich \\
    Zurich, Switzerland
}

\usepackage{xcolor}
\newif\ifbitlogicdraft
\bitlogicdraftfalse
\ifbitlogicdraft

  \newcommand{\phfig}[1]{\fbox{\parbox{0.9\linewidth}{\centering\textcolor{red!70!black}{\small\textbf{FIGURE PLACEHOLDER.} #1}}}}
\else

  \newcommand{\phfig}[1]{}
\fi
\def\month{06}  
\def\year{2026} 
\def\openreview{\url{https://openreview.net/forum?id=ZbsSZAfDod}} 

\begin{document}

\maketitle

\begin{abstract}
Gradient-based LUT- and logic-gate-based neural networks (LUTNet,
LogicNets, DiffLogic, PolyLUT, NeuraLUT, WARP-LUT, DWN, LILogicNet,
LightLUT) replace multiply-accumulate arithmetic with Boolean
lookups. The same trained checkpoint deploys to GPU as bitwise ops on
bit-packed activations, to FPGA as LUT primitives, and to ASIC as
standard-cell gates, all from one code path. Yet each method ships
its own training pipeline, encoder, connectivity rule, fan-in, and
hardware-reporting convention. The natural practitioner question,
which of these choices actually matter for accuracy and which for
hardware cost, therefore has no answer in the current literature. We
release \textbf{BitLogic}, a unified framework that factors the field into a
five-axis design space (encoder, connectivity, fan-in, node
parameterization, head) and instantiates every prior method under one
shared training and evaluation protocol. The framework deliberately
omits method-specific procedures such as calibration, pruning, and
thresholding, and all evaluations are limited to two-layer
feed-forward networks. Combining the per-axis
winners identifies a new best-of-space configuration that outperforms
every retrained prior on every (dataset, width) cell in which every
compared prior fits the shared budget, across MNIST, Fashion-MNIST,
CIFAR-10, and CIFAR-100. We evaluate the best-of-space model on all
three backends. On MNIST the resulting two-layer network reaches
${\sim}126$\,MSamples/s on FPGA, ${\sim}15\times$ the throughput of a
bit-packed GPU forward path that itself processes $64$ samples per
$64$-bit operation, at four-to-five orders of magnitude less energy.
\end{abstract}
\acresetall

\section{Introduction}
\label{sec:introduction}

Machine-learning inference now dominates the energy footprint of
deployed models. \citet{yang2024doubleexponentialincreasesinferenceenergy}
report that machine-learning workloads accounted for 10 to 15\% of
Google's total energy use between 2019 and 2021, roughly 60\% of which
was inference, and Meta reports a 10:20:70 split across experimentation,
training, and inference. This has motivated a line of
\ac{LUT}-native networks
\citep{wang2019lutnetrethinkinginferencefpga,umuroglu2020logicnetscodesignedneuralnetworks,petersen2022deepdifferentiablelogicgate,andronic2025polylutultralowlatencypolynomial,andronic2025neuralutassemblehardwareawareassemblingsubneural,gerlach2025warplutswalshassistedrelaxation,bacellar2025differentiableweightlessneuralnetworks,fojcik2025lilogicnetcompactlogic,ruttgers2025lightdifferentiablelogicgate}
in which each neuron is a small \ac{LUT} or Boolean gate. The same
trained model runs as bitwise ops on a bit-packed \ac{GPU} forward
path, maps directly onto \ac{FPGA} \ac{LUT} primitives, and synthesizes
to standard-cell \ac{ASIC} gates, so a single checkpoint covers three
deployment backends without retraining.

\paragraph{The fragmented design space.}
The nine gradient-based \ac{LUT} methods we know of (LUTNet, LogicNets,
DiffLogic, PolyLUT, NeuraLUT, WARP-LUT, \ac{DWN}, LILogicNet, LightLUT)
share the same goal but pick different encoders, connectivity rules,
fan-ins, truth-table relaxations, output heads, and hardware-reporting
conventions. Their published numbers are therefore not directly
comparable. Which choices matter for accuracy and which for hardware
cost, and whether any published method sits near the best point of
the shared design space, are open questions in the current literature.

\paragraph{Our approach.}
We factor the union of all nine methods into five independent design
axes: input encoder, per-layer connection map, per-node Boolean fan-in,
node parameterization, and output head. Every prior method becomes one
point in this space (\Cref{sec:design_space},
\Cref{tab:prior_mapping}). We implement all five axes in
\textbf{BitLogic} as independently swappable, then sweep each axis one
at a time on MNIST under a single shared training protocol
(\Cref{sec:protocol}). Combining the per-axis winners identifies a new
best-of-space configuration that no prior paper has trained
(\Cref{tab:prior_mapping}, last row).

We then retrain the six gradient-trained priors with a published
weight-learning recipe (DiffLogic, PolyLUT, NeuraLUT, \ac{DWN},
WARP-LUT, LILogicNet) inside the framework under the same protocol,
on MNIST, Fashion-MNIST, CIFAR-10, and CIFAR-100
(\Cref{sec:benchmark}). LUTNet, LogicNets, and LightLUT are placed in
the same five-axis map but not independently retrained. Where the
shared protocol allows, the retrained columns reproduce each method's
reported accuracy. Where it does not, the gap is traceable to
method-specific machinery (PolyLUT's structured pruning, \ac{DWN}'s
calibration, WARP-LUT's residual block) that the shared protocol
disables on purpose. Finally, the best-of-space model deploys to three
backends from one code path: a bit-packed \ac{GPU} path, Vivado
post-route on two Xilinx \acp{FPGA}, and a target-independent
\ac{ASIC} proxy via Yosys and Nangate $45$\,nm
(\Cref{sec:hardware_evaluation}).

\paragraph{Contributions.}
\begin{itemize}
\itemsep0em
\item The BitLogic framework, released so subsequent work can
evaluate a new axis against the whole published slate with one
command, with all five axes (encoder, connectivity, fan-in, node
parameterization, head) independently swappable
(\Cref{sec:design_space}).
\item A protocol-matched cross-method comparison: per-axis sweeps
that identify a new best-of-space model (a combination no prior paper
has trained, \Cref{sec:sweep}), and six gradient-trained priors
retrained on four standard image-classification benchmarks
(\Cref{sec:benchmark}).
\item A unified \ac{GPU}$\,+\,$\ac{FPGA}$\,+\,$\ac{ASIC} evaluation
of the resulting checkpoint from one code path, with Python and
deployed hardware accuracy bit-exact by construction
(\Cref{sec:hardware_evaluation}).
\end{itemize}

\paragraph{Scope.}
Empirical claims are limited to feedforward two-layer image
classification. Convolutional, residual, attention, and recurrent
extensions are deliberately out of scope. The two-layer restriction
and the open depth axis are revisited in \Cref{sec:discussion}.

\section{Related work}
\label{sec:related_work}

\paragraph{Weightless neural networks.}
Learning over lookup tables and bit-level logic predates the recent
\ac{FPGA}-deployment wave: \mbox{n-tuple} networks
\citep{bledsoe1959ntuple} and \ac{WiSARD} \citep{aleksander1984wisard}
store Boolean responses in RAM cells addressed by small input tuples.
BTHOWeN~\citep{susskind2022bthowen} bridges the classical weightless
line to the modern \ac{FPGA}-deployment wave with Bloom-filter RAM
neurons and a hardware-accelerator design, and its extended
finite-difference gradient estimator is the direct ancestor of the
\ac{DWN} parameterization studied below. Modern differentiable-\ac{LUT}
methods revisit this idea with gradient-based training.

\paragraph{Gradient-based LUT methods.}
Starting with LUTNet \citep{wang2019lutnetrethinkinginferencefpga} and
LogicNets \citep{umuroglu2020logicnetscodesignedneuralnetworks}, a
sequence of methods has proposed different continuous relaxations of
Boolean \ac{LUT} functions, trainable by gradient descent and
discretized afterward. Differentiable logic-gate networks
\citep{petersen2022deepdifferentiablelogicgate} relax each node as a
softmax over the $2^{2^K}$ Boolean functions of fan-in $K$; the
parameter count scales double-exponentially in $K$, so this work is
limited to $K{=}2$ in practice. LightLUT
\citep{ruttgers2025lightdifferentiablelogicgate} hits the same node
family with a lean Kronecker-indicator basis and soft or
straight-through-hard forward sampling, dropping the per-node parameter
count to $2^K$. BitLogic instantiates this parametrization at
$K\in\{2,4,6\}$ and evaluates it head-to-head with the relaxations
below. LILogicNet \citep{fojcik2025lilogicnetcompactlogic} adds
\emph{learnable Top-$k$ sparse routing} on top of the same node family,
selecting each node's $K$ inputs from a small candidate pool through a
differentiable mask. Other \ac{LUT}-native relaxations keep the fan-in
larger by structuring the truth table: PolyLUT
\citep{andronic2023polylutlearningpiecewise,
andronic2025polylutultralowlatencypolynomial} fits a piecewise
polynomial and bakes it into the \ac{LUT}, NeuraLUT
\citep{andronic2024neuralut,
andronic2025neuralutassemblehardwareawareassemblingsubneural} absorbs
a small dense sub-network into each \ac{LUT}, WARP-LUT
\citep{gerlach2025warplutswalshassistedrelaxation} reparameterizes in
the Walsh basis, and \ac{DWN}
\citep{bacellar2025differentiableweightlessneuralnetworks} keeps the
full $2^n$-entry truth table and trains it with an extended
finite-difference gradient estimator. The design-space coordinates of
the six retrained methods (DiffLogic, PolyLUT, NeuraLUT, \ac{DWN},
WARP-LUT, LILogicNet) appear in \Cref{tab:prior_mapping}.

\paragraph{Orthogonal and complementary directions.}
Arithmetic \ac{FPGA} accelerators such as HLS4ML
\citep{fast_duarte_2018} and FINN / FINN-R
\citep{Umuroglu_2017,blott2018finnr} target quantized \ac{MAC}
pipelines on \ac{DSP} slices, and LUTMUL \citep{xie2025lutmul} re-hosts
\ac{MAC} on \acp{LUT}; these are complementary to the \ac{LUT}-native
setting of this paper because the underlying primitive is still
arithmetic. A survey of \ac{LUT}-based \ac{FPGA} \acp{DNN} is
\citet{guo2025surveylutbaseddeepneural}. Topology-focused approaches
that vary convolutional and interconnect structure rather than the
node itself
\citep{petersen2024convolutionaldifferentiablelogicgate,
kresse2025scalableinterconnectlearningboolean}, hybrid architectures
that combine \ac{LUT} neurons with conventional arithmetic layers
\citep{nag2025llvitedgedeployablevision}, and \ac{LUT}-based models
that avoid continuous relaxation entirely such as TreeLUT
\citep{Khataei_2025} are orthogonal to the node-parameterization
comparison of this paper and represent natural extensions of the
design-space view.

\section{The five-axis design space}
\label{sec:design_space}

A feedforward \ac{LUT} network is the composition of five
interchangeable components:
\begin{equation}
\label{eq:design_space}
\underbrace{\mathcal{E}}_{\text{encoder}}
\;\to\;
\Bigl[\,\underbrace{L_{\mathcal{M}}(f_{\vtheta}^{(n)})}_{\text{layer: connectivity $\mathcal{M}$, nodes of fan-in $n$ and parameterization $f_{\vtheta}$}}\,\Bigr]^{D}
\;\to\;
\underbrace{\mathcal{H}}_{\text{head}}.
\end{equation}
\Cref{eq:design_space} is read left to right as the forward
composition of a single network: a real input vector of dimension $d$
(the number of input features, e.g.\ $784$ for flattened MNIST) is
first mapped by the encoder $\mathcal{E}$ to $d b$ binary wires, where
$b$ is the number of bits emitted per input dimension; the result is
then passed through $D$ stacked Boolean logic layers (the
$[\,\cdot\,]^{D}$ bracket denotes $D$ repetitions) and finally
aggregated by the head $\mathcal{H}$ into $c$ real-valued class
scores. The five axes are (i) the encoder
$\mathcal{E}: \R^{d} \to \{0,1\}^{d b}$, (ii) the per-layer connection
map $\mathcal{M}$, (iii) the per-node Boolean fan-in $n$, (iv) the node
parameterization $f_{\vtheta}$, and (v) the head
$\mathcal{H}: \{0,1\}^{w} \to \R^{c}$. Depth $D$ and per-layer width
are architecture-level knobs layered on top. Every prior method in
\Cref{tab:prior_mapping} fixes a particular setting of these axes.
BitLogic exposes all five as independently configurable.

\paragraph{Nodes are parameterizations, not paradigms.}
Each layer holds $w$ Boolean nodes, each implementing an $n$-input
Boolean function. Methods differ only in how they relax that discrete
object to a differentiable surrogate $f_{\vtheta}$ during training.
The per-method choice is summarized in \Cref{tab:prior_mapping}. At
inference every surrogate discretizes to the same $2^{2^n}$ Boolean
truth table, so the deployed \ac{LUT} format is identical across the
field. What differs is which subset of that space is \emph{reachable
during training}. PolyLUT at degree $d$ only spans the
$\binom{n}{\leq d}$ Fourier subspace, while NeuraLUT is capped by the
per-neuron \ac{MLP}'s capacity, until the \ac{LUT}-extraction step
rounds each to a full truth table. Parameterizations therefore differ
in optimization dynamics and in the corner of the Boolean hypothesis
space each can approach, not in the deployed format.

\paragraph{Connectivity is the main per-layer design choice.}
A layer with fan-in $n$ and output width $w$ selects an input tuple
per node through a per-node connection mapping
$\mathcal{M}_j: \{1,\dots,n\} \to \{1,\dots,w_{\text{in}}\}$, where
$j \in \{1,\dots,w\}$ indexes the output node (one map per node) and
$w_{\text{in}}$ is the layer's input width (the encoder output width
$d b$ for the first layer, the previous layer's $w$ otherwise). Each
$\mathcal{M}_j$ assigns the node's $n$ Boolean inputs to wires of the
layer input. Two
families are common: \emph{fixed} routing, where $\mathcal{M}$ is
frozen at construction (random or random-unique initialization) and
only the node parameters are learned, and \emph{learnable} routing,
where the selection matrix is differentiable during training and
snapped to a valid sparse mapping at discretization. The learnable
family is further parameterized by a per-slot candidate-pool size,
from four candidates up to the full input width (the full-width
setting collapses to a matmul fast-path).

\paragraph{Fan-in is the dominant hardware knob.}
The hardware cost of a discrete $n$-input \ac{LUT} grows as $O(2^n)$
in the worst case. Larger $n$ gives each node more expressive power
but more than doubles its \ac{LUT} footprint. We therefore treat
fan-in as a first-class axis of the comparison rather than a
node-specific detail.

\paragraph{Encoders and heads.}
The encoder $\mathcal{E}$ maps each continuous input dimension to $b$
bits. We evaluate three families. The \emph{linear thermometer}
\citep{buckman2018thermometer} uses equispaced thresholds, while the
\emph{distributive (quantile) thermometer}
\citep{bacellar2022distributivethermometer} uses empirical
training-set quantiles. Both emit $b{+}1$ distinct levels on $b$
wires. The \emph{uniform fixed-point} encoder is a Brevitas-style
narrow-range integer quantizer that emits $2^b$ levels on $b$ wires
and is the representation used by LogicNets, PolyLUT, and NeuraLUT.
The head $\mathcal{H}$ aggregates the final binary feature vector into
$c$ real-valued class logits. We use the popcount head studied by
\citet{brandle2025mnisttoimagenet} alongside a quantized variant: a
$\tanh$-bounded $c{\times}c$ weight matrix snapped to a signed integer
grid that maps onto DSP48 multipliers.

\section{Experimental protocol}
\label{sec:protocol}

All experiments share one training and evaluation recipe. This single
recipe is used throughout, so the cross-method
comparison of \Cref{sec:benchmark} controls for everything except each
method's design-space coordinates (\Cref{sec:design_space},
\Cref{tab:prior_mapping}) and any structural cap it imposes.

\paragraph{Architecture.}
A two-layer feedforward \ac{LUT} network (encoder $\to$ logic layer
$\to$ logic layer $\to$ head), per \Cref{eq:design_space}. Per-layer
width is reported per-experiment. Encoder, connectivity, fan-in, node
parameterization, and head are the five swept axes and are specified
per table.

\paragraph{Training.}
AdamW ($\beta_1{=}0.9$, $\beta_2{=}0.999$, $\epsilon{=}10^{-8}$),
constant learning rate $\eta{=}0.01$ with no schedule or warmup, batch
size $128$, weight decay $0$, no label smoothing, no gradient clipping,
no mixed precision. Loss is cross-entropy on the raw head logits. Every
run trains for $100$ epochs with early stopping disabled, two seeds
per cell, mean $\pm$ std reported. Per-parameterization initialization,
augmentation policy, and seed control are documented in
\Cref{sec:appendix_reproducibility}.

\paragraph{Data.}
MNIST, Fashion-MNIST, CIFAR-10, and CIFAR-100 from torchvision. CIFAR
training splits are augmented (random horizontal flip, random crop with
reflect-mode padding). MNIST and Fashion-MNIST are not. The native
training split is divided 90/10 into training and validation via a
seed-controlled permutation. The test split is the native torchvision
test set. No channel-wise normalization, cutout, or mixup.

\paragraph{Evaluation.}
Every accuracy number we report comes from a bit-packed \ac{GPU}
inference path that evaluates the discretized \ac{LUT}s with bitwise
operations. This path is bit-exact with the emitted SystemVerilog by
construction (\Cref{sec:appendix_hw_eval}), so every accuracy number
is simultaneously a deployment accuracy. The full hardware pipeline
(HDL emission, Vivado post-route, Yosys $+$ Nangate $45$\,nm) and the
definition of each hardware-table column are in
\Cref{sec:appendix_hw_eval}.

\section{Design-space sweep: finding a new best model}
\label{sec:sweep}

\newcommand{\experitable}{%
  \footnotesize
  \setlength{\tabcolsep}{4.5pt}%
  \renewcommand{\arraystretch}{1.15}%
}

We sweep each of the five design-space axes of \Cref{sec:design_space}
one at a time on MNIST, at three widths per sweep so axis signal is
separable from width scaling (full ladder rationale in
\Cref{sec:appendix_width_ladders}). Every sweep varies one axis at a
time on top of a common base architecture: LightLUT (soft) nodes,
learnable routing with eight candidates per slot, a quantile
thermometer at $b{=}8$, fan-in $n{=}4$, and a popcount head. All
runs follow the shared protocol of \Cref{sec:protocol}.

\subsection{Node parameterization}
\label{sec:exp_node}

We sweep the seven node parameterizations of \Cref{sec:design_space},
listing LightLUT with soft and with straight-through-hard
forward sampling as two rows. The gap between the two rows measures
the soft-to-hard discretization gap that
\citet{yousefi2025mindgapremovingdiscretization} analyze in detail
for differentiable logic-gate networks. All rows use $n{=}4$ except
DiffLogic, which is locked to $n{=}2$ by construction.

\begin{table}[htb]
\centering
\experitable
\caption{Node parameterization sweep on MNIST. Accuracy is mean $\pm$
std over two seeds at each width. \textbf{NAND2-GE} and \textbf{LUTs}
report the standard-cell and Vivado-post-route cost of the
\texttt{layers} submodule at the narrowest width. Smaller is better,
bold per column. LightLUT (soft) is the per-width accuracy winner.
The top cluster (LightLUT, WarpLUT, DwnLUT) lies within
${\sim}0.7$\,pp at $w{\leq}16$K and tightens to ${\sim}0.3$\,pp at
$w{=}32$K, and most of the DiffLogic deficit is a fan-in effect, not a
node-parameterization one.}
\label{tab:rq_node}
\begin{tabular}{@{} l r r r r r @{}}
\toprule
\textbf{Node} & \textbf{Acc @ 8\,K} & \textbf{Acc @ 16\,K} & \textbf{Acc @ 32\,K} & \textbf{NAND2-GE @ 8\,K} & \textbf{LUTs @ 8\,K} \\
\midrule
\texttt{LightLUT} (soft) & $\mathbf{91.14 \pm 0.02}$ & $\mathbf{94.82 \pm 0.02}$ & $\mathbf{97.11 \pm 0.05}$ & $60{,}649 \pm 164$ & $15{,}242 \pm 20$ \\
\texttt{LightLUT} (hard) & $91.04 \pm 0.06$ & $94.62 \pm 0.10$ & $96.94 \pm 0.04$ & $61{,}600 \pm 195$ & $15{,}284 \pm 17$ \\
\texttt{WarpLUT} & $91.03 \pm 0.05$ & $94.67 \pm 0.03$ & $96.88 \pm 0.05$ & $65{,}542 \pm 138$ & $15{,}314 \pm 4$ \\
\texttt{PolyLUT} & $90.97 \pm 0.04$ & $94.09 \pm 0.00$ & $96.53 \pm 0.09$ & $37{,}264 \pm 257$ & $12{,}358 \pm 31$ \\
\texttt{DwnLUT} & $90.77 \pm 0.02$ & $94.50 \pm 0.07$ & $96.94 \pm 0.01$ & $60{,}869 \pm 179$ & $15{,}223 \pm 26$ \\
\texttt{NeuraLUT} & $90.63 \pm 0.05$ & $93.73 \pm 0.19$ & $96.22 \pm 0.07$ & $34{,}956 \pm 459$ & $12{,}064 \pm 103$ \\
\texttt{LinearLUT} & $90.17 \pm 0.04$ & $93.39 \pm 0.10$ & $95.84 \pm 0.05$ & $33{,}626 \pm 13$ & $11{,}404 \pm 4$ \\
\texttt{DiffLogicLUT} ($n{=}2$) & $84.78 \pm 0.13$ & $89.21 \pm 0.03$ & $93.53 \pm 0.06$ & $\mathbf{12{,}441 \pm 164}$ & $\mathbf{7{,}138 \pm 4}$ \\
\bottomrule
\end{tabular}

\end{table}

\begin{figure}[htb]
\centering
\begin{subfigure}[t]{0.48\linewidth}
    \centering
    \includegraphics[width=\linewidth]{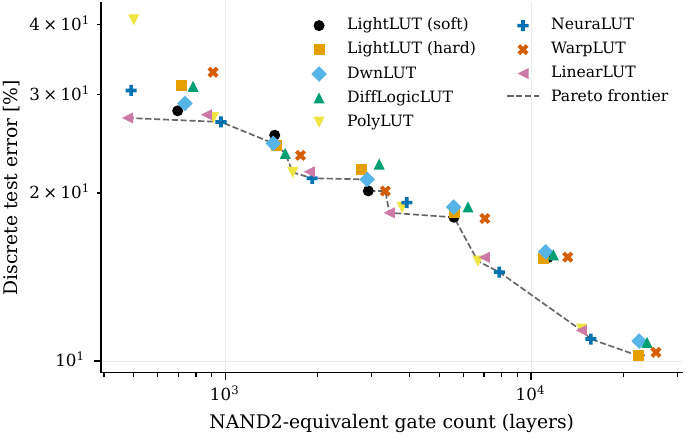}
    \caption{Cost on the \texttt{layers} submodule as standard-cell
    NAND2-equivalent gate count (Yosys $+$ Nangate $45$\,nm).}
    \label{fig:node_rank2_nand}
\end{subfigure}\hfill
\begin{subfigure}[t]{0.48\linewidth}
    \centering
    \includegraphics[width=\linewidth]{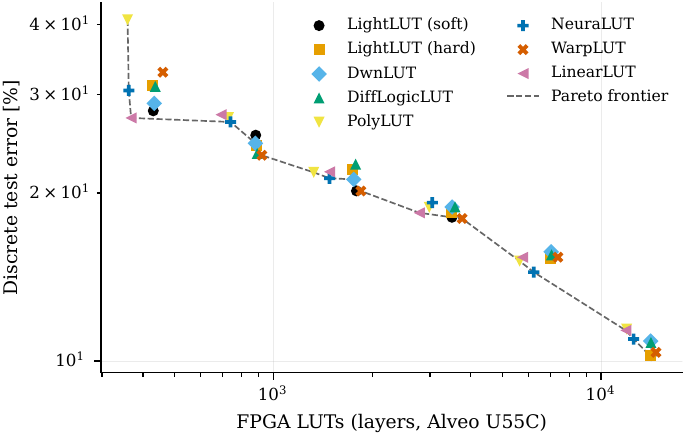}
    \caption{Cost as Vivado post-route LUT count of the
    \texttt{layers} submodule on the Alveo U55C target.}
    \label{fig:node_rank2_lut}
\end{subfigure}
\caption{MNIST discrete test error versus hardware cost at matched
$n{=}2$ / $b{=}4$, one series per node parameterization, swept over
six widths, both axes log-scaled. Error is evaluated in the
HDL-matching forward mode (\Cref{sec:appendix_hw_emit}), so the
reported value equals the deployed-network error rate.}
\end{figure}

\newpage
\Cref{fig:node_rank2_nand,fig:node_rank2_lut} project the same axis
onto the cost dimension at matched $n{=}2$ / $b{=}4$: each node
parameterization is retrained across six widths
$w \in \{500,\,1{,}000,\,2{,}000,\,4{,}000,\,8{,}000,\,16{,}000\}$ and
plotted against a Yosys $+$ Nangate $45$\,nm NAND2-equivalent gate
count (\Cref{fig:node_rank2_nand}) and against a Vivado post-route
\ac{LUT} count on the Alveo U55C (\Cref{fig:node_rank2_lut}).

At $w{=}32$K the top cluster of \Cref{tab:rq_node}
(LightLUT (soft/hard), WarpLUT, DwnLUT) sits
within ${\sim}0.3$\,pp of each other ($96.88$--$97.11$\,\%), with
PolyLUT and NeuraLUT trailing by
${\sim}0.6$--$0.9$\,pp. Under matched $n{=}2$ in
\Cref{fig:node_rank2_nand} the DiffLogic gap to the $n{=}4$
cluster (of ${\sim}3.6$\,pp in \Cref{tab:rq_node}) narrows to
${\sim}2$\,pp, so most of the DiffLogic deficit is a fan-in
effect rather than a node-parameterization effect. The per-family
error spread narrows with width (from ${\sim}1.5\times$ at the cheap
end of the cost sweep to ${\sim}1.1\times$ at the wide end), so the
choice of node parameterization matters most under tight hardware
budgets.

\newpage

\subsection{Connectivity}
\label{sec:exp_conn}

We sweep six connectivity configurations (\Cref{tab:rq_conn}): fixed
routing with random or random-unique initialization, and learnable
routing at candidate-pool sizes
$k \in \{4,\,8,\,16,\,\text{full layer}\}$. The
ladder is $w \in \{1{,}000,\,2{,}000,\,4{,}000\}$, narrower than the
base ladder because the full-layer learnable cell does not fit at
$w{=}8\text{K}$ (\Cref{sec:appendix_width_ladders}).

\begin{table}[htb]
\centering
\experitable
\caption{Connectivity sweep on MNIST at $w \in \{1{,}000,\,2{,}000,\,4{,}000\}$.
All other axes inherit the base architecture. NAND2-GE and \acp{LUT}
are reported on the \texttt{layers} submodule at the narrowest width.
Smaller is better, bold per column. Bounded candidate pools
($k{=}8$--$16$) maximize accuracy. Fixed (random-unique) routing
trails the best learnable variant by only ${\sim}0.9$\,pp. Full-layer
learnable routing mode-collapses $7$--$8$\,pp below every bounded
variant. Its NAND2-GE minimum is the hardware shadow of that collapse
(see prose and \Cref{sec:appendix_cost_model}), not an efficiency
gain.}
\label{tab:rq_conn}
\begin{tabular}{@{} l r r r r r @{}}
\toprule
\textbf{Connectivity} & \textbf{Acc @ 1\,K} & \textbf{Acc @ 2\,K} & \textbf{Acc @ 4\,K} & \textbf{NAND2-GE @ 1\,K} & \textbf{LUTs @ 1\,K} \\
\midrule
Learnable, full layer (\texttt{-1}) & $79.10 \pm 0.88$ & $81.18 \pm 0.92$ & $81.47 \pm 0.18$ & $\mathbf{3{,}908 \pm 4}$ & $\mathbf{1{,}262 \pm 4}$ \\
Learnable, $16$ candidates & $\mathbf{86.19 \pm 0.27}$ & $\mathbf{87.28 \pm 0.13}$ & $\mathbf{88.84 \pm 0.06}$ & $7{,}325 \pm 212$ & $1{,}850 \pm 5$ \\
Learnable, $8$ candidates & $85.02 \pm 0.45$ & $87.22 \pm 0.09$ & $88.51 \pm 0.01$ & $7{,}117 \pm 39$ & $1{,}828 \pm 10$ \\
Learnable, $4$ candidates & $85.38 \pm 0.29$ & $86.90 \pm 0.01$ & $88.55 \pm 0.23$ & $7{,}347 \pm 41$ & $1{,}852 \pm 4$ \\
Fixed (random) & $84.55 \pm 0.20$ & $86.95 \pm 0.11$ & $88.19 \pm 0.42$ & $7{,}620 \pm 66$ & $1{,}896 \pm 2$ \\
Fixed (random-unique) & $84.10 \pm 0.10$ & $86.62 \pm 0.16$ & $87.97 \pm 0.01$ & $7{,}755 \pm 29$ & $1{,}917 \pm 9$ \\
\bottomrule
\end{tabular}

\end{table}

The full-layer learnable cell is the anomaly of this sweep. It drops
to $79.10\%$ at $w{=}1$K and $81.47\%$ at $w{=}4$K, $7$--$8$\,pp
below every bounded-candidate variant. With no candidate-pool
bottleneck the softmax router mode-collapses: many slots converge on
the same high-signal wires (a ${\sim}20\%$ reduction in distinct
input wires per layer between the $k{=}16$ and full-pool
configurations, measured from the end-of-training argmax routing).
When multiple \ac{LUT} inputs tie to the same wire, the $16$-entry
truth table reduces to an effective $\leq$$3$-input function and
Yosys collapses the gate network accordingly, so the NAND2-GE
reduction on that row is a hardware shadow of the same pathology
visible in accuracy, not an efficiency gain. Among the
non-pathological rows, fixed (random-unique) routing is within
${\sim}0.9$\,pp of the best learnable variant at $w{=}4$K with no
candidate-selection overhead.

\subsection{Fan-in}
\label{sec:exp_fanin}

We sweep $n \in \{2,\,4,\,6\}$ across the three-width ladder with all
other axes at the base architecture (\Cref{tab:rq_fanin}). Fan-in
controls the per-node
truth-table size ($2^n$ entries) and is the axis where the standard-cell
cost grows fastest, so this sweep maps the accuracy / cost knee that
pins the cross-method comparison at $n{=}4$
(\Cref{sec:appendix_reproducibility}).

\begin{table}[htb]
\centering
\experitable
\caption{Fan-in sweep on MNIST at three widths. NAND2-GE and LUTs
are reported on the \texttt{layers} submodule at the narrowest width.
 Accuracy grows roughly logarithmically with $n$ and
the $n{=}2{\to}4$ step is the largest. The standard-cell cost of
$n{=}6$ is ${\sim}4\times$ that of $n{=}4$. On LUT fabrics $n \leq 6$
maps into a single 6-input LUT, so the ASIC knee is real while the
LUT-fabric knee is nearly absent.}
\label{tab:rq_fanin}
\begin{tabular}{@{} l r r r r r @{}}
\toprule
\textbf{Fan-in} & \textbf{Acc @ 8\,K} & \textbf{Acc @ 16\,K} & \textbf{Acc @ 32\,K} & \textbf{NAND2-GE @ 8\,K} & \textbf{LUTs @ 8\,K} \\
\midrule
$n{=}6$ & $\mathbf{93.83 \pm 0.05}$ & $\mathbf{96.44 \pm 0.01}$ & $\mathbf{97.61 \pm 0.00}$ & $248{,}588 \pm 779$ & $16{,}134$ \\
$n{=}4$ & $91.13 \pm 0.02$ & $94.85 \pm 0.03$ & $97.12 \pm 0.03$ & $60{,}673 \pm 125$ & $15{,}250 \pm 14$ \\
$n{=}2$ & $84.69 \pm 0.08$ & $89.58 \pm 0.11$ & $93.75 \pm 0.06$ & $\mathbf{11{,}358 \pm 192}$ & $\mathbf{7{,}085 \pm 35}$ \\
\bottomrule
\end{tabular}

\end{table}

\subsection{Encoder}
\label{sec:exp_encoder}

We sweep three encoder families, each parameterized by a bit width
$b$ (\Cref{tab:rq_encoder}). The uniform and quantile
(``distributive'') thermometers are
evaluated at $b \in \{4,\,8\}$. The binary-coded uniform quantizer at
$b \in \{2,\,8\}$. The low-bit point of the binary code isolates the
regime where it outperforms the thermometers at half the wire count.
$b{=}8$ is common across all three for a matched high-bit comparison.

\begin{table}[htb]
\centering
\experitable
\caption{Encoder sweep on MNIST across three widths. All other axes
inherit the base architecture. NAND2-GE and \acp{LUT} are reported on
the \texttt{encoder} submodule at the narrowest width. Smaller is
better, bold per column. The single ``---'' cell marks a synthesis
job that exceeded its memory budget and was not retried.
 At matched nominal bit widths the quantile
thermometer matches or beats the uniform one by ${\sim}0.5$\,pp on
MNIST, but most of that gain and essentially all of the NAND2-GE
reduction at $b{=}4$ is an MNIST-specific threshold collapse
(\Cref{sec:appendix_encoder_collapse}). The binary-coded quantizer at
$b{=}2$ halves the wire count but costs ${\sim}4\times$ more NAND2-GE
than the distributive thermometer at $b{=}4$.}
\label{tab:rq_encoder}
\begin{tabular}{@{} l r r r r r @{}}
\toprule
\textbf{Encoder (bits)} & \textbf{Acc @ 8\,K} & \textbf{Acc @ 16\,K} & \textbf{Acc @ 32\,K} & \textbf{NAND2-GE @ 8\,K} & \textbf{LUTs @ 8\,K} \\
\midrule
\texttt{DistributiveThermometer}, $b{=}4$ & $\mathbf{91.26 \pm 0.01}$ & $\mathbf{94.84 \pm 0.07}$ & $\mathbf{97.15 \pm 0.03}$ & $\mathbf{3{,}397}$ & $\mathbf{1{,}568}$ \\
\texttt{DistributiveThermometer}, $b{=}8$ & $91.15 \pm 0.00$ & $\mathbf{94.84 \pm 0.03}$ & $97.09 \pm 0.01$ & $7{,}690$ & $3{,}136$ \\
\texttt{Thermometer}, $b{=}4$ & $90.39 \pm 0.05$ & $94.19 \pm 0.12$ & $96.89 \pm 0.09$ & $14{,}112$ & $5{,}488$ \\
\texttt{Thermometer}, $b{=}8$ & $90.42 \pm 0.17$ & $94.27 \pm 0.08$ & $96.81 \pm 0.09$ & $35{,}221$ & $10{,}976$ \\
\texttt{UniformFixedPoint}, $b{=}2$ & $89.78 \pm 0.07$ & $93.51 \pm 0.04$ & $96.38 \pm 0.02$ & $13{,}034$ & $3{,}920$ \\
\texttt{UniformFixedPoint}, $b{=}8$ & $88.59 \pm 0.05$ & $92.61 \pm 0.10$ & $95.55 \pm 0.04$ & --- & $12{,}544$ \\
\bottomrule
\end{tabular}

\end{table}

At matched nominal bit widths the quantile thermometer matches or
beats the linear one by ${\sim}0.5$\,pp on MNIST (consistent with
\citet{bacellar2022distributivethermometer}). Most of the gain and
essentially all of the NAND2-GE reduction at $b{=}4$ is an
MNIST-specific threshold collapse rather than a real encoder win
(\Cref{sec:appendix_encoder_collapse}).

\subsection{Head}
\label{sec:exp_head}

We compare the popcount head with a \ac{DSP}-backed quantized head
(a $c{\times}c$ signed matmul on the length-$c$ group-sum vector with
$\text{wbits}{=}8$, so weights are $\tanh$-bounded during training
and snapped to the signed integer grid $[-127,\,127]$ at inference,
matching the integer MAC the emitter drops into the HDL) across the
three-width ladder (\Cref{tab:rq_head}).

\begin{table}[htb]
\centering
\experitable
\caption{Head sweep on MNIST across three widths. All other axes
inherit the base architecture. NAND2-GE, \acp{LUT}, and \acp{DSP} are
reported on the \texttt{head} submodule at the narrowest width.
 The popcount head is competitive at $w{=}32$K
($-0.7$\,pp) with zero \acp{DSP}. The \ac{DSP}-backed head buys
$+0.7$--$4.6$\,pp across the width ladder at $39$ \acp{DSP},
${\sim}10\%$ more FPGA \acp{LUT}, and ${\sim}28\%$ more
NAND2-GE: attractive when \acp{DSP} are abundant and width is
constrained, essentially unjustified at $w{=}32$K.}
\label{tab:rq_head}
\begin{tabular}{@{} l r r r r r r @{}}
\toprule
\textbf{Head} & \textbf{Acc @ 8\,K} & \textbf{Acc @ 16\,K} & \textbf{Acc @ 32\,K} & \textbf{NAND2-GE @ 8\,K} & \textbf{LUTs @ 8\,K} & \textbf{DSPs @ 8\,K} \\
\midrule
\texttt{GroupedDSP} & $\mathbf{95.77 \pm 0.33}$ & $\mathbf{97.42 \pm 0.08}$ & $\mathbf{97.81 \pm 0.08}$ & $75{,}666$ & $12{,}028 \pm 224$ & $39 \pm 1$ \\
\texttt{GroupSum} & $91.14 \pm 0.00$ & $94.84 \pm 0.02$ & $97.12 \pm 0.04$ & $\mathbf{59{,}080}$ & $\mathbf{10{,}883}$ & $\mathbf{0}$ \\
\bottomrule
\end{tabular}

\end{table}

\FloatBarrier
\section{Cross-method benchmark}
\label{sec:benchmark}

We retrain the six gradient-trained priors (DiffLogic, PolyLUT,
NeuraLUT, \ac{DWN}, WARP-LUT, LILogicNet) inside the framework under
the shared protocol of \Cref{sec:protocol}, each mapped via
\Cref{tab:prior_mapping}. To keep the comparison fair, the BitLogic
best-of-space row pins fan-in to $n{=}4$ and uses the popcount head
across all rows. The rationale (and the remaining axis values from
\Cref{sec:sweep}) is in \Cref{sec:appendix_reproducibility}.

\begin{table}[!htbp]
\centering
\experitable
\begin{threeparttable}
\caption{Best-of-design-space BitLogic against six gradient-trained
prior methods retrained inside BitLogic under the common protocol.
Accuracy (\%) is mean $\pm$ std over two seeds at each width, except
the WARP-LUT CIFAR-100 $w{=}64$K cell, which is single-seed
(reported without $\pm$ for that reason). ``---''
in the Published column means the original paper does not report that
(method, dataset) combination. \ac{DWN} \textsc{oom} cells are
infrastructure-limited by its $O(w^2)$ fast-path and $10$-bit
thermometer rather than a statement about the method itself
(note [f]). Gaps between the Retrained and Published columns reflect
the method-specific pipelines the shared protocol disables
(\Cref{sec:appendix_reproducibility}), not the design-space
coordinates.}
\label{tab:best_of_space}
\begin{tabular}{@{} l l r r r r @{}}
\toprule
\textbf{Dataset} & \textbf{Method} & \textbf{Acc @ 4\,K} & \textbf{Acc @ 16\,K} & \textbf{Acc @ 64\,K} & \textbf{Published} \\
\midrule
\multirow{7}{*}{MNIST}
 & DWN        & $87.11 \pm 0.34$ & \textsc{oom}\tnote{f} & \textsc{oom}\tnote{f} & 97.80--98.77\tnote{d} \\ 
 & PolyLUT    & $86.24 \pm 0.09$ & $91.61 \pm 0.17$ & $96.22 \pm 0.00$ & 96.0--97.5\tnote{b} \\ 
 & NeuraLUT   & $85.91 \pm 0.33$ & $91.53 \pm 0.11$ & $96.03 \pm 0.09$ & 96.0\tnote{c} \\ 
 & LILogicNet & $80.40 \pm 0.05$ & $88.37 \pm 0.16$ & $95.73 \pm 0.08$ & 97.96--98.95\tnote{e} \\ 
 & WARP-LUT   & $76.40 \pm 0.22$ & $85.89 \pm 0.10$ & $93.59 \pm 0.04$ & --- \\ 
 & DiffLogic  & $75.92 \pm 0.30$ & $85.89 \pm 0.03$ & $93.68 \pm 0.05$ & 97.69--98.47\tnote{a} \\ 
 & \textbf{BitLogic best-of-space} & $\mathbf{88.94 \pm 0.14}$ & $\mathbf{95.05 \pm 0.07}$ & $\mathbf{97.84 \pm 0.04}$ & --- \\
\midrule
\multirow{7}{*}{F-MNIST}
 & DWN        & $78.09 \pm 0.15$ & \textsc{oom}\tnote{f} & \textsc{oom}\tnote{f} & 89.01--89.12\tnote{d} \\ 
 & PolyLUT    & $74.65 \pm 0.24$ & $83.08 \pm 0.28$ & $87.00 \pm 0.12$ & --- \\ 
 & NeuraLUT   & $74.27 \pm 0.08$ & $82.46 \pm 0.03$ & $86.48 \pm 0.13$ & --- \\ 
 & LILogicNet & $66.50 \pm 0.25$ & $75.54 \pm 0.07$ & $82.10 \pm 0.01$ & --- \\ 
 & WARP-LUT   & $65.65 \pm 0.38$ & $75.36 \pm 0.12$ & $83.00 \pm 0.05$ & --- \\ 
 & DiffLogic  & $65.42 \pm 0.05$ & $75.40 \pm 0.04$ & $82.89 \pm 0.12$ & 87.44\tnote{d} \\ 
 & \textbf{BitLogic best-of-space} & $\mathbf{78.38 \pm 0.14}$ & $\mathbf{85.90 \pm 0.22}$ & $\mathbf{89.16 \pm 0.08}$ & --- \\
\midrule
\multirow{7}{*}{CIFAR-10}
 & PolyLUT    & $36.95 \pm 0.02$ & $46.16 \pm 0.01$ & $53.02 \pm 0.16$ & --- \\ 
 & NeuraLUT   & $36.73 \pm 0.02$ & $44.38 \pm 0.33$ & $47.30 \pm 0.65$ & --- \\ 
 & WARP-LUT   & $33.86 \pm 0.10$ & $42.92 \pm 0.29$ & $52.12 \pm 0.01$ & --- \\ 
 & LILogicNet & $33.83 \pm 0.13$ & $42.36 \pm 0.20$ & $51.67 \pm 0.09$ & 55.11--60.98\tnote{e} \\ 
 & DiffLogic  & $33.72 \pm 0.16$ & $42.55 \pm 0.02$ & $51.73 \pm 0.34$ & 51.27--62.14\tnote{a} \\ 
 & DWN        & \textsc{oom}\tnote{f} & \textsc{oom}\tnote{f} & \textsc{oom}\tnote{f} & 57.42--57.51\tnote{d} \\ 
 & \textbf{BitLogic best-of-space} & $\mathbf{38.93 \pm 0.19}$ & $\mathbf{49.22 \pm 0.26}$ & $\mathbf{58.06 \pm 0.14}$ & --- \\
\midrule
\multirow{7}{*}{CIFAR-100}
 & PolyLUT    & $8.79 \pm 0.46$ & $12.02 \pm 0.15$ & $16.10 \pm 0.02$ & --- \\ 
 & NeuraLUT   & $8.71 \pm 0.22$ & $11.88 \pm 0.04$ & $15.20 \pm 0.17$ & --- \\ 
 & LILogicNet & $7.63 \pm 0.01$ & $10.62 \pm 0.12$ & $14.54 \pm 0.04$ & --- \\ 
 & DiffLogic  & $7.49 \pm 0.21$ & $10.61 \pm 0.08$ & $14.64 \pm 0.09$ & --- \\ 
 & WARP-LUT   & $7.00 \pm 0.04$ & $10.46 \pm 0.00$ & $14.43$ & --- \\ 
 & DWN        & \textsc{oom}\tnote{f} & \textsc{oom}\tnote{f} & \textsc{oom}\tnote{f} & --- \\ 
 & \textbf{BitLogic best-of-space} & $\mathbf{10.19 \pm 0.06}$ & $\mathbf{14.06 \pm 0.04}$ & $\mathbf{18.82 \pm 0.09}$ & --- \\ 
\bottomrule
\end{tabular}

\begin{tablenotes}[flushleft]\footnotesize
\item[] Published numbers are reproduced from the cited works and are
not protocol-matched. They reflect each paper's own training budget
and hardware conventions.
\item[a] DiffLogic: \citet{petersen2022deepdifferentiablelogicgate}.
\item[b] PolyLUT: \citet{andronic2023polylutlearningpiecewise,
andronic2025polylutultralowlatencypolynomial}.
\item[c] NeuraLUT: \citet{andronic2024neuralut}; the published cell
quotes only the 2024 NeuraLUT recipe, which is the recipe the retrained
row reproduces. NeuraLUT-Assemble 2025
\citep{andronic2025neuralutassemblehardwareawareassemblingsubneural}
reports a higher MNIST accuracy of $98.6\%$ but is a successor recipe
not reproduced under the shared protocol
(\Cref{sec:appendix_reproducibility}).
\item[d] \ac{DWN}: \citet{bacellar2025differentiableweightlessneuralnetworks}.
The F-MNIST DiffLogic entry is a cross-fill from \ac{DWN}'s Table~1,
since F-MNIST is absent from DiffLogic's own paper.
\item[e] LILogicNet: \citet{fojcik2025lilogicnetcompactlogic}.
\item[f] \ac{DWN}'s LR-\ac{DWN} dense connectome uses the full-layer
learnable fast-path, whose $O(w^2)$ routing matrix exceeds the training
\ac{GPU} memory budget beyond $w{\approx}4{,}000$; only the $w{=}4{,}000$
cell is retrained, and the 10-bit thermometer input on
CIFAR-10 / CIFAR-100 pushes even that out of budget. LILogicNet is
retrained at its paper's flagship Top-$32$ connectivity and fits the
full ladder.
\end{tablenotes}
\end{threeparttable}
\end{table}

Under the shared protocol, the best-of-space configuration wins every
retrained (dataset, width) cell in which every compared prior fits the
training budget: $62$ of the $72$ ($4$ datasets ${\times}\,3$ widths
${\times}\,6$ priors) cells in \Cref{tab:best_of_space}; the remaining
$10$ are \ac{DWN} \textsc{oom} entries (note [f]) and are not part of
that count. At $w{=}64$K the best-of-space reaches
$97.84 / 89.16 / 58.06 / 18.82$\,\% on
MNIST / F-MNIST / CIFAR-10 / CIFAR-100; the two largest margins are
$\sim$1.6\,pp over PolyLUT on MNIST and $\sim$5\,pp over PolyLUT on
CIFAR-10.

Where the published row exceeds the retrained row by more than
${\sim}1$\,pp (\ac{DWN} on MNIST: $97.80$--$98.77$ vs.\ retrained
$87.11$; LILogicNet on MNIST: $97.96$--$98.95$ vs.\ retrained
$95.73$ at $w{=}64$K; NeuraLUT-Assemble on MNIST, $98.6$, not
retrained), the gap traces to a method-specific calibration,
pruning, or thresholding pipeline that the shared protocol disables
(full list in \Cref{sec:appendix_reproducibility}). Under the shared
protocol, the design-space coordinates of the BitLogic best-of-space
dominate every retrained prior at every width that fits the training
budget.

\section{Hardware deployment}
\label{sec:hardware_evaluation}

One checkpoint emits to three backends from a single framework: a
bit-packed \ac{GPU} forward path that processes $64$ samples per
$64$-bit operation, synthesizable SystemVerilog placed and routed by
Vivado on two \ac{FPGA} targets (Alveo U55C, Zynq UltraScale+
XCZU7EV), and a flat Yosys netlist against Nangate $45$\,nm as a
target-independent \ac{ASIC} proxy. The \ac{GPU} deployment runs on
three consumer cards (\Cref{sec:appendix_hw_eval}). We deploy the MNIST winner of
\Cref{tab:best_of_space} at $w{=}4{,}000$, the widest cell our build
host fits. Wider cells extrapolate linearly at unchanged timing
(\Cref{sec:appendix_hw_width}). \ac{FPGA} numbers are Vivado
post-route, with Verilator confirming bit-exact agreement against the
Python reference on the full MNIST test set.

\paragraph{Emission modes.}
Each \ac{FPGA} row picks one of three modes: \textbf{max throughput}
(parallel encoder/head, fully pipelined, $\mathrm{II}{=}1$),
\textbf{lowest latency} (parallel encoder/head, no pipelining), and
\textbf{fewest resources} (parallel encoder, combinational layers,
iterative popcount head over $\mathrm{II}{=}c$ cycles). On the MNIST
winner only head iteration pays back. The per-knob breakdown is in
\Cref{sec:appendix_cost_model}.

\paragraph{Cross-platform headline.}
\Cref{tab:hardware_evaluation} reports $F_\text{max}$, latency,
throughput, power, and energy per sample across the three \ac{FPGA}
modes, both targets, and three \ac{GPU} references at two batch sizes.
All rows share the same checkpoint, so accuracy is identical
($88.79\%$ on the full MNIST test set, within the two-seed
$88.94\pm0.14\%$ band of \Cref{tab:best_of_space}). Resource
footprints are in \Cref{tab:hardware_resources_fpga,tab:hardware_resources_asic,tab:hardware_resources_gpu}.

\begin{table}[!htbp]
\centering
\experitable
\caption{Speed and power profile of the MNIST winner at $88.79\%$ test
accuracy (single-seed $w{=}4{,}000$ checkpoint): three \ac{FPGA}
emission modes on two targets plus three \ac{GPU} references.
\textbf{Lat (ns) mixes per-batch and per-sample}: \ac{GPU} rows report
batch wall-clock, \ac{FPGA} rows per-sample pipeline latency at steady
state. Throughput is comparable across platforms, latency is not.
Post-route \ac{FPGA} estimates, not on-board. The bottom row group
quotes two published \ac{FPGA}
baselines~\citep{blott2018finnr,borras2022opensourcefpgamlcodesign}.
They are not directly comparable to our rows. Provenance and caveats
are in \Cref{sec:appendix_hw_external}. Methodology details in
\Cref{sec:appendix_hw_eval}.}
\label{tab:hardware_evaluation}
\adjustbox{max width=\textwidth}{%
\begin{tabular}{@{} l l r r r r r @{}}
\toprule
\textbf{Platform} & \textbf{Mode} & \textbf{$F_\text{max}$} & \textbf{Lat} & \textbf{Tput} & \textbf{Power} & \textbf{Energy} \\
 &  & \small(MHz) & \small(ns) & \small(kSmp/s) & \small(W) & \small(nJ/smp) \\
\midrule
\multirow{2}{*}{RTX 3090} & batchsize 64 & $2{,}115$ & $115{,}712$ & $550.0$ & $130.67$ & $2{,}287{,}786$ \\
 & batchsize 1024 & $2{,}115$ & $116{,}736$ & $8663.7$ & $291.42$ & $319{,}425$ \\
\midrule
\multirow{2}{*}{RTX 2080 Ti} & batchsize 64 & $2{,}100$ & $114{,}224$ & $553.5$ & $58.02$ & $1{,}014{,}773$ \\
 & batchsize 1024 & $2{,}100$ & $146{,}656$ & $6984.6$ & $199.60$ & $227{,}080$ \\
\midrule
\multirow{2}{*}{TITAN RTX} & batchsize 64 & $2{,}100$ & $118{,}784$ & $533.0$ & $72.49$ & $1{,}273{,}177$ \\
 & batchsize 1024 & $2{,}100$ & $135{,}088$ & $7587.7$ & $218.91$ & $246{,}347$ \\
\midrule
\multirow{3}{*}{U55C} & max throughput & $126.6$ & $40$ & $126566.3$ & $3.40$ & $27$ \\
 & lowest latency & $84.4$ & $12$ & $84359.7$ & $3.34$ & $40$ \\
 & fewest resources & $110.8$ & $99$ & $10069.7$ & $3.37$ & $335$ \\
\midrule
\multirow{3}{*}{ZU7EV} & max throughput & $127.2$ & $39$ & $127177.9$ & $0.72$ & $6$ \\
 & lowest latency & $84.7$ & $12$ & $84652.5$ & $0.66$ & $8$ \\
 & fewest resources & $115.8$ & $95$ & $10528.0$ & $0.68$ & $65$ \\
\midrule
\multicolumn{7}{@{}l}{\textit{External references}} \\
\midrule
Ultra96 (ZUS+ ZCU3EG) & FINN-R MLP-4 (MNIST)~\citep{blott2018finnr} & $300.0$ & --- & $851.7$ & $11.80$ & $13{,}856$ \\
PYNQ-Z1 (Zynq-7020) & FINN-R MLP-4 (MNIST)~\citep{blott2018finnr} & $100.0$ & --- & $162.3$ & $2.50$ & $15{,}400$ \\
Pynq-Z2 (Zynq-7020) & ResNet-V1 hls4ml (CIFAR-10)~\citep{borras2022opensourcefpgamlcodesign} & --- & $27{,}300{,}000$ & --- & --- & $44{,}330{,}000$ \\
Pynq-Z2 (Zynq-7020) & CNV-W1A1 FINN (CIFAR-10)~\citep{borras2022opensourcefpgamlcodesign} & --- & $1{,}500{,}000$ & --- & --- & $2{,}535{,}000$ \\
\bottomrule
\end{tabular}
}
\end{table}

The three modes span a clear resource--throughput trade-off:
max-throughput buys ${\sim}1.5\times$ throughput for three orders of
magnitude more flip-flops, while \emph{fewest resources} cuts
\acp{LUT} by ${\sim}1.5\times$. The pipelined implementation reaches
$126.6$\,MSamp/s (U55C) and $127.2$\,MSamp/s (XCZU7EV), about
$15\times$ the RTX\,3090 and $17$--$18\times$ the Turing cards, at
$6$--$27$\,nJ per sample, four to five orders of magnitude below
\ac{GPU}. The published \ac{FPGA} numbers in the bottom row group of
\Cref{tab:hardware_evaluation} are not a fair comparison, see
\Cref{sec:appendix_hw_external}.

\FloatBarrier

\section{Discussion, limitations, and outlook}
\label{sec:discussion}

\paragraph{What the sweep taught us.}
Two axes give clean signal and two surface structural surprises.
\emph{Fan-in} is the single largest lever: the $n{=}2{\to}4$ step
alone explains most of DiffLogic's gap to the $n{=}4$ cluster, and
the matched-$n{=}2$ figures (\Cref{fig:node_rank2_nand,fig:node_rank2_lut})
narrow that gap from ${\sim}3.6$\,pp to ${\sim}2$\,pp. Once $n$ is
fixed, the top \emph{node cluster} (LightLUT, WarpLUT, DwnLUT)
converges within ${\sim}0.3$\,pp at $w{=}32$K
(\Cref{tab:rq_node}), so the relaxation family matters much less than
the literature suggests. \emph{Full-layer learnable connectivity}
mode-collapses $7$--$8$\,pp below every bounded-candidate variant
(\Cref{sec:exp_conn}). The apparent NAND2-GE win on that row is a
hardware shadow of the pathology, not an efficiency gain. The
\emph{quantile thermometer}'s lead on MNIST is largely an artefact
of an MNIST-specific threshold collapse (\Cref{sec:appendix_encoder_collapse}),
not encoder dominance: $(\text{pixel}>0)$ already carries most of the
MNIST signal.

\paragraph{Where the parameterization win shows up.}
The per-axis ranking of \Cref{tab:rq_node} compares node
parameterizations at fixed width, and the best-of-space combination
of \Cref{tab:best_of_space} amplifies that ranking. On \ac{GPU} this
translates directly into runtime, since cost is set by parameter
count and width. \Cref{fig:node_rank2_nand,fig:node_rank2_lut}
re-project the same axis against \ac{ASIC} NAND2-equivalent gate
count and \ac{FPGA} \ac{LUT} count, and the per-family error spread
shrinks from about $1.5\times$ at the cheap end of the cost sweep to
about $1.1\times$ at the wide end. On those fabric backends the
node-parameterization choice is therefore largely orthogonal: any of
the node parameterizations sits within about $1.1\times$ of any
other at deployment-relevant cost. The accuracy headroom from
picking the best-of-space combination matters most when parameter
count drives runtime cost, that is, on \ac{GPU}.

\paragraph{Depth, the open sixth axis.}
Every configuration we evaluate is two layers deep, and we deliberately did
not sweep depth. Randomly-initialized \ac{LUT} stacks lose accuracy
rapidly with depth (pilot runs show the soft-vs-hard training gap
widening and gradients vanishing through stacked saturating
relaxations), and every known remedy introduces a new axis: residual
identity initialization
\citep{petersen2024convolutionaldifferentiablelogicgate} and
fixed-connectivity routing both mitigate the collapse. We flag depth
as the next unsolved axis rather than the next parameter.

\paragraph{Extensibility.}
The five axes are independently swappable at the layer level: a
reader with a new node, encoder, or connectivity rule can rerun
\Cref{sec:sweep} with one config change and measure the new axis
against the whole retrained slate. Whole-model topology is fixed in
the current release (dense logic layers plus a small set of heads and
encoders), but the extension points are clear: convolutional logic
layers \citep{petersen2024convolutionaldifferentiablelogicgate},
residual/skip variants, and recurrent \ac{LUT} layers
\citep{buhrer2025recurrentdeepdifferentiablelogic} all exist in
isolation. The natural next step is to make each a first-class axis
so the same protocol-matched comparison applies. The framework is not
bound to image classification, and extending it to other tasks is
feasible.

\paragraph{Outlook.}
The pipelined \ac{FPGA} deployment of \Cref{sec:hardware_evaluation}
already runs at sustained throughput high enough that on the edge the
sensor and downstream pipeline, not the network, become the
bottleneck. Whether this profile extends to attention, residual, or
recurrent \ac{LUT}-native networks at \ac{LLM} scale is the open
question BitLogic is meant to enable: every modern neural-network
success story (attention, residual learning, convolution, recurrence)
has been written in the floating-point idiom, and each needs a
\ac{LUT}-native counterpart before the edge and serving efficiencies
reported here transfer to serving-scale workloads. The five axes we
swept are the feedforward base case. The next axes are topology,
depth, and task.

\section{Conclusion}
\label{sec:conclusion}

Casting every gradient-based feedforward \ac{LUT} method as a point
in one five-axis design space and sweeping each axis under a shared
protocol yields a new best-of-space model, a combination no prior
paper has trained, that matches or exceeds every retrained baseline
on $62$ of the $72$ (dataset, width, prior) cells of
\Cref{tab:best_of_space} across MNIST, Fashion-MNIST, CIFAR-10, and
CIFAR-100. The same checkpoint, emitted to bit-exact SystemVerilog
and to a standard-cell netlist, reaches the throughput and energy
figures reported in \Cref{sec:hardware_evaluation}. While this
best-of-space accuracy headroom translates into a visible \ac{GPU}
runtime advantage, on \ac{FPGA} and \ac{ASIC} the cost-versus-accuracy
curves of \Cref{fig:node_rank2_nand,fig:node_rank2_lut} cluster across
all node parameterizations, so the parameterization choice mainly
matters for \ac{GPU} efficiency.

\ac{LUT}-based networks push inference onto extremely low-energy
hardware, which is broadly positive for the energy footprint of
deployed machine-learning systems. Extending the framework beyond
image classification is a natural next step.

\bibliography{main}
\bibliographystyle{tmlr}

\clearpage
\appendix
\section{Reproducibility: initialization, prior-method mapping, sweep recipes}
\label{sec:appendix_reproducibility}

This appendix supplements the training protocol of
\Cref{sec:protocol} with initialization detail, the mapping of each
prior method onto the five design-space axes used for the retrained
rows of \Cref{tab:best_of_space}, the axis constraints that pin those
BitLogic rows at $n{=}4$ and the popcount head, the fidelity scope of
the retrained column, the per-experiment sweep recipes, and the
training-cost numbers. The hardware pipeline (bit-packed inference,
HDL emission, Vivado post-route flow, Yosys $+$ Nangate $45$\,nm
standard-cell synthesis) is deferred to \Cref{sec:appendix_hw_eval}.

\paragraph{Initialization.}
Fixed routing draws its index tensor once per layer as either
i.i.d.\ random or random-unique. Learnable routing draws the
candidate-pool mask uniformly and initializes the per-slot routing
logits to zero, so the expected selection is uniform across
candidates at the first step. Node parameters are drawn from a
zero-mean Gaussian (every parametrization's
\texttt{weight\_init="random"} default). The library also exposes
an opt-in residual-anchor init that biases each node toward
pass-through of its last input, but it is not enabled in any sweep
reported here. Quantile-thermometer thresholds are fit once from the
empirical pixel quantiles on the training split. The linear
thermometer uses equispaced thresholds in the value range.

\paragraph{Seeds.}
Seeds $\in \{0, 1\}$ drive torch, NumPy, and the 90/10
train/validation split. Every per-axis and cross-method experiment
reports mean $\pm$ std over both seeds. The rank-2 node sweep
(\Cref{fig:node_rank2_nand,fig:node_rank2_lut}) is single-seed by
design: it is a cost-vs-accuracy projection rather than a mean
comparison, and the within-seed cost spread is already dominated by
the width sweep itself.

\paragraph{Prior-method mapping.}
Each prior method in \Cref{tab:best_of_space} corresponds to the
design-space coordinates listed in \Cref{tab:prior_mapping} and is
trained at three widths $w \in \{4{,}000,\,16{,}000,\,64{,}000\}$ on
each of the four datasets. Two exceptions: \ac{DWN}'s LR-DWN dense
connectome uses the full-layer learnable fast-path, whose $O(w^2)$
routing matrix exceeds the \ac{GPU} memory budget beyond
$w{\approx}4{,}000$, so \ac{DWN} is retrained at $w{=}4{,}000$ only
on MNIST and Fashion-MNIST and is \textsc{oom} at every width on
CIFAR-10 / CIFAR-100 (where its $10$-bit thermometer further inflates
the input tensor). LILogicNet is retrained at its paper's flagship
Top-$32$ connectivity (the 2Top32 variant,
\citealp{fojcik2025lilogicnetcompactlogic} Table~3), whose
$O(w{\cdot}32)$ routing matrix fits the full ladder. ``Learnable($k$)''
denotes learnable routing with candidate-pool size $k$, and $k{=}{-}1$
denotes the full-layer fast-path. Encoder family and per-dataset bit
count $b$ come from each method's paper. Missing datasets use a
nearest-neighbour extrapolation (grayscale $\leftrightarrow$ MNIST,
RGB $\leftrightarrow$ CIFAR-10). For DiffLogic, $b{=}1$ on MNIST
reproduces the paper's binarized-MNIST single threshold at $0.5$, and
$b{=}3$ on CIFAR-10 reproduces the ``small'' row of their Table~5
(fixed per-channel thresholds $\{0.25, 0.5, 0.75\}$). The
$31$-threshold ``large'' CIFAR-10 variant is not retrained.

\paragraph{Axis constraints in the cross-method comparison.}
Two of the five axes are held fixed across all BitLogic rows of
\Cref{tab:best_of_space} rather than taken from the per-axis winners
of \Cref{sec:sweep}. \textbf{Fan-in is pinned to $n{=}4$} because
(i) the majority of retrained priors (PolyLUT, NeuraLUT, and \ac{DWN})
natively evaluate at rank $4$, and (ii) the fan-in sweep
(\Cref{tab:rq_fanin}) shows accuracy improves faster with width than
with rank on the prior-method ladder, so $n{=}4$ keeps most priors at
their native rank while retaining width headroom. DiffLogic, WARP-LUT,
and LILogicNet, which default to $n{=}2$ in their own papers, therefore
sit at rank $4$ here as a controlled design-space coordinate, not an
accidental mismatch. \textbf{The head is pinned to the popcount
GroupSum} so the comparison stays pure \ac{LUT}-only: a \ac{DSP}-backed
head would add a quantized matrix-multiply that no prior method uses,
turning the table into a hybrid \ac{LUT}$+$\ac{DSP} vs.\ \ac{LUT}-only
contest. The remaining three axes (node parameterization, connectivity,
encoder) are set to the per-axis winners of
\Cref{tab:rq_node,tab:rq_conn,tab:rq_encoder}.

\paragraph{Scope of the prior-method reproduction.}
\Cref{tab:prior_mapping} aligns each retrained baseline with its paper
along the five design-space axes. This is an axis-level alignment,
not a full re-implementation. The protocol of \Cref{sec:protocol} is shared
across all rows, so the comparison is controlled for everything except
the design-space choices, and method-specific machinery is intentionally
out of scope. PolyLUT-2025's hardware-aware structured-pruning pipeline
(dense pre-train with exponential-$\ell_1$ group regularizer, top-$k$
prune, retrain from reinitialized weights,
\citealp{andronic2025polylutultralowlatencypolynomial}) and
NeuraLUT-Assemble's successor recipe
\citep{andronic2025neuralutassemblehardwareawareassemblingsubneural}
are not reproduced, so the PolyLUT and NeuraLUT rows use the
fixed (random) sparsity of the 2023 / 2024 originals rather than
learned connectivity. \ac{DWN}'s bespoke schedule and quantile-thermometer
calibration \citep{bacellar2025differentiableweightlessneuralnetworks}
are replaced by the shared recipe. LILogicNet's flagship Top-$32$
learnable connectivity \citep{fojcik2025lilogicnetcompactlogic} is kept
verbatim, but its dataset-specific binarization thresholds are replaced
by linear thermometer thresholds at $i/(b{+}1)$. WARP-LUT's full
residual convolutional block
\citep{gerlach2025warplutswalshassistedrelaxation} is replaced by the
flat dense backbone used by every other row. The resulting numbers
should therefore be read as ``each method's parameterization and
connection scheme, trained under BitLogic's uniform recipe'', not as a
reproduction of each headline number.

\begin{table}[htb]
\centering
\experitable
\caption{Prior-method to design-space mapping for the retrained rows
of \Cref{tab:best_of_space}. $n$ is the per-node fan-in, $b$ is the
number of thresholds per input feature or colour channel. Superscript
$\dagger$ marks a dataset not evaluated in the paper, the value is our
closest-analogue extrapolation. The head axis is pinned to the
popcount GroupSum across every row (see the axis-constraints
paragraph above). The final row is the BitLogic best-of-space
configuration identified by the per-axis winners of
\Cref{sec:sweep}, which is trained as a new point in the design
space rather than imported from any prior method.}
\label{tab:prior_mapping}
\adjustbox{max width=\textwidth}{%
\begin{tabular}{@{} l l l c l c c c c @{}}
\toprule
 &  &  &  &  & \multicolumn{4}{c}{\textbf{Encoder bits $b$}} \\
\cmidrule(lr){6-9}
\textbf{Method} & \textbf{Node parameterization} & \textbf{Connections} & \textbf{$n$} & \textbf{Encoder} & MNIST & F-MNIST & CIFAR-10 & CIFAR-100 \\
\midrule
DiffLogic    & softmax over 16 ops         & fixed (random)     & 2 & linear thermometer    & 1 & $1^\dagger$ & 3 & $3^\dagger$ \\
PolyLUT      & degree-$d$ polynomial       & fixed (random)     & 4 & binary-coded quantizer & 2 & $2^\dagger$ & $2^\dagger$ & $2^\dagger$ \\
NeuraLUT     & MLP per neuron              & fixed (random)     & 4 & binary-coded quantizer & 2 & $2^\dagger$ & $2^\dagger$ & $2^\dagger$ \\
DWN          & full truth table (EFD)      & learnable($-1$)    & 6 & quantile thermometer   & 3 & 7          & 10         & $10^\dagger$ \\
WARP-LUT     & Walsh basis                 & fixed (random)     & 2 & linear thermometer    & $1^\dagger$ & $1^\dagger$ & 3 & $3^\dagger$ \\
LILogicNet   & softmax over 16 ops         & learnable($32$)    & 2 & linear thermometer    & 1 & $1^\dagger$ & 7 & $7^\dagger$ \\
\midrule
\textbf{BitLogic best-of-space} & LightLUT (soft) & learnable($16$) & 4 & quantile thermometer & 4 & 4 & 4 & 4 \\
\bottomrule
\end{tabular}%
}
\end{table}

\paragraph{Per-experiment recipes.}
\Cref{tab:exp_recipes} lists each paper element with the axis it
sweeps, the swept values, and the width ladder. Standard-cell
synthesis is not invoked for the per-axis sweeps or the cross-method
comparison. Hardware cost is reserved for the rank-2 node figures
(\Cref{fig:node_rank2_nand,fig:node_rank2_lut}) and the deployment
case study (\Cref{sec:hardware_evaluation}).

\begin{table}[htb]
\centering
\experitable
\caption{Per-experiment sweep configurations behind the tables and
figures of \Cref{sec:sweep,sec:benchmark}. Unless stated otherwise
the swept axis is the only deviation from the base architecture of
\Cref{sec:sweep} (LightLUT (soft), learnable($8$), $n{=}4$,
$b{=}8$ quantile thermometer, popcount head). Every configuration is
trained with two seeds. Accuracy is read from the bit-packed
inference path on the native test split.}
\label{tab:exp_recipes}
\begin{tabular}{@{} l p{0.55\linewidth} l @{}}
\toprule
\textbf{Element} & \textbf{Swept axis and values} & \textbf{Width ladder} \\
\midrule
\Cref{tab:rq_node}        & node: \{LightLUT (soft, hard), DwnLUT, PolyLUT, NeuraLUT, WarpLUT, LinearLUT, DiffLogic\} & $\{8,16,32\}$\,K \\
\Cref{tab:rq_conn}        & connectivity: \{fixed (random), fixed (random-unique), learnable($k$) for $k{\in}\{4,8,16,-1\}$\} & $\{1,2,4\}$\,K \\
\Cref{tab:rq_fanin}       & fan-in: $n \in \{2, 4, 6\}$ & $\{8,16,32\}$\,K \\
\Cref{tab:rq_encoder}     & encoder $\times$ bits: \{uniform, quantile\}$\times$\{4,8\}, binary$\times$\{2,8\} & $\{8,16,32\}$\,K \\
\Cref{tab:rq_head}        & head: \{popcount, DSP-backed ($\text{wbits}{=}8$)\} & $\{8,16,32\}$\,K \\
\Cref{tab:best_of_space}  & BitLogic best + 6 prior methods (\Cref{tab:prior_mapping}), $\times$ 4 datasets & $\{4,16,64\}$\,K \\
\Cref{fig:node_rank2_nand,fig:node_rank2_lut} & node sweep at rank-2 / $b{=}4$ / learnable($8$) / popcount, single seed & $\{0.5,1,2,4,8,16\}$\,K \\
\Cref{tab:hardware_evaluation}--\Cref{tab:hardware_resources_asic} & deployment case study driven by the MNIST winner of \Cref{tab:best_of_space} & $w{=}4$\,K \\
\bottomrule
\end{tabular}
\end{table}

The BitLogic cells in \Cref{tab:best_of_space} use the best-of-space
configuration identified by the per-axis winners of
\Cref{sec:sweep}: LightLUT (soft) nodes, learnable routing with
$k{=}16$ candidates, $n{=}4$, a $b{=}4$ distributive thermometer, and
a popcount head.

\paragraph{Training cost.}
\Cref{tab:training_cost_details} reports per-parameterization
wall-clock and peak \ac{GPU} memory at the largest node-sweep width
($w{=}32{,}000$). Training runs on a shared heterogeneous \ac{GPU}
cluster whose scheduler allocates whichever device is free, so seeds
of the same cell routinely land on different silicon. Seconds per
epoch scale roughly linearly with width. Within each row the two
seeds are grouped by host \ac{GPU} rather than averaged across
dissimilar cards.

\begin{table}[htb]
\centering
\experitable
\caption{Training wall-clock, peak GPU memory, and
trainable-parameter count per node parameterization at the largest
width in the node sweep ($w{=}32{,}000$, 2-layer MNIST base). Rows are
grouped by host \ac{GPU} (``/''-separated) rather than averaged across
dissimilar silicon. \textit{s / epoch}, \textit{Total (min)}, and
\textit{Peak GPU (GB)} mirror that split in the same order, one
sub-cell per \ac{GPU}. Parameter counts are identical across seeds at
a fixed configuration.}
\label{tab:training_cost_details}
\begin{tabular}{@{} l l r r r r @{}}
\toprule
\textbf{Parametrization} & \textbf{GPU host(s)} & \textbf{Trainable params} & \textbf{s / epoch} & \textbf{Total (min)} & \textbf{Peak GPU (GB)} \\
\midrule
\texttt{LightLUT} (hard) & RTX 3090 / TITAN RTX & $3{,}072{,}000$ & $35.50$ / $34.07$ & $59.16$ / $56.78$ & $2.535$ / $2.535$ \\
\texttt{LightLUT} (soft) & RTX 3090 & $3{,}072{,}000$ & $26.93$ & $44.89$ & $2.535$ \\
\texttt{DwnLUT} & RTX 3090 / TITAN RTX & $3{,}072{,}000$ & $49.31$ / $26.90$ & $82.18$ / $44.83$ & $1.806$ / $1.806$ \\
\texttt{DiffLogicLUT} ($n{=}2$) & RTX 3090 / TITAN RTX & $2{,}048{,}000$ & $\mathbf{15.04}$ / $19.51$ & $\mathbf{25.07}$ / $32.52$ & $\mathbf{0.936}$ / $\mathbf{0.936}$ \\
\texttt{PolyLUT} & RTX 2080 Ti & $4{,}288{,}000$ & $50.28$ & $83.80$ & $3.320$ \\
\texttt{NeuraLUT} & RTX 3090 / TITAN RTX & $5{,}184{,}000$ & $42.60$ / $36.63$ & $71.00$ / $61.05$ & $2.220$ / $2.220$ \\
\texttt{WarpLUT} & RTX 2080 Ti & $3{,}072{,}000$ & $44.11$ & $73.51$ & $2.451$ \\
\texttt{LinearLUT} & RTX 3090 & $2{,}368{,}000$ & $37.47$ & $62.45$ & $1.852$ \\
\bottomrule
\end{tabular}

\end{table}

\section{Extended sweep data}
\label{sec:appendix_extended_sweeps}

This appendix collects the experimental detail that is orthogonal to
the narrative of \Cref{sec:sweep,sec:benchmark,sec:hardware_evaluation}
but that the reader may need when interrogating the headline numbers:
why each sweep uses the width ladder it does, which axes are held fixed
in the cross-method comparison and why, and which emitter knobs pay
back on the MNIST winner in \Cref{sec:hardware_evaluation}.

\subsection{Width ladders}
\label{sec:appendix_width_ladders}

The per-axis sweeps are reported at three widths per axis so that axis
signal is separable from width scaling. The node, fan-in, encoder, and
head sweeps of \Cref{sec:sweep} use the base ladder
$w \in \{8{,}000,\,16{,}000,\,32{,}000\}$, centred on the $w{=}16{,}000$
base cell. The connectivity sweep uses the smaller ladder
$w \in \{1{,}000,\,2{,}000,\,4{,}000\}$ because its full-layer
learnable cell materializes a dense $w{\times}w$ routing matrix per
layer and does not fit at the 8\,K base within our training \ac{GPU}
memory budget. The relative ordering of the bounded-candidate rows
at that ladder carries over to the wider ladder when re-run, but the
full-layer cell is not reachable above $w{\approx}4{,}000$ under this
recipe. The cross-method comparison of \Cref{tab:best_of_space} uses
the wider ladder $w \in \{4{,}000,\,16{,}000,\,64{,}000\}$ for every
row that fits, so the deployment regime is visible. \ac{DWN}'s
LR-DWN dense connectome inherits the same full-layer fast-path
and is retrained at $w{=}4{,}000$ only, with wider cells marked
\textsc{oom} in the same sense.

\subsection{Cost-model asymmetry of the FPGA emitter}
\label{sec:appendix_cost_model}

The \emph{fewest resources} design point of \Cref{sec:hardware_evaluation}
is a selected combination of the emitter knobs, not a blanket ``iterate
everything'' switch. Each knob was measured on the MNIST winner and
kept only when it actually reduces whole-model cost.

\paragraph{Head iteration wins on both cost models.}
Time-sharing the per-class popcount and argmax reduces the head from
${\sim}5{,}170$ \acp{LUT} (${\sim}29.5$\,k NAND2-eq) to ${\sim}2{,}120$
\acp{LUT} (${\sim}7.6$\,k NAND2-eq) for $k{=}10$ classes. The per-class
reduction is expensive enough that time-sharing it still wins after
paying for the counter and output-latch flip-flops.

\paragraph{Encoder iteration loses at small input scales.}
The parallel encoder for the MNIST winner is already cheap:
$s{\cdot}b = 784{\cdot}4 = 3{,}136$ thermometer bits, $3{,}397$
NAND2-eq total, because each channel is a short comparator chain.
Replacing it with a time-shared comparator over $784$ slots requires a
$3{,}136$-bit output register whose standard-cell cost alone
(${\sim}15{,}680$ NAND2-eq at ${\sim}5$ NAND2 per flip-flop) dominates
the combinational savings, and iterative emission measures
${\sim}5.8\times$ worse end-to-end. On \ac{FPGA} the flip-flops pack
into slice positions already bundled with the \acp{LUT}, so the
register cost is near-zero in slice-\ac{LUT} accounting, but the
input-gather multiplexer still drives the encoder \ac{LUT} count up
from $1{,}568$ to $3{,}301$. At larger input scales the ratio flips,
and the MNIST winner is simply too encoder-small for iteration to
pay back.

\paragraph{Layer iteration loses on random-wiring LUT fabrics.}
Each \ac{LUT}'s $n$ inputs are independently learned indices into the
prior layer's output (no spatial locality, no convolutional structure),
so a time-shared layer fabric needs a per-neuron $w{:}1$ multiplexer
on each of its $n$ inputs to switch between the two layers' id
tensors. At $w{=}4{,}000$ and $n{=}4$ that gather cost is comparable
to the \ac{LUT}-lookup itself, and adding a $4{,}000$-bit inter-layer
register tips the balance. Layer emission therefore stays
combinational.

The selected combination (parallel encoder, combinational layers,
iterative head) minimizes both \ac{FPGA} \ac{LUT} count and
Nangate\,45\,nm gate count on this model. The asymmetry is intrinsic
to the interaction between emitter knobs and cost model, not a defect
of any one knob. A bigger encoder or a structured connectome would
flip these verdicts.

\subsection{MNIST-specific encoder collapse}
\label{sec:appendix_encoder_collapse}

The NAND2-eq column of \Cref{tab:rq_encoder} needs a caveat: the
DistributiveThermometer row synthesizes to a quarter of the
Thermometer row at matched nominal bit widths (${\sim}3{,}400$
vs.\ ${\sim}14{,}100$ at $b{=}4$), but these two encoders are not
operating at matched \emph{effective} bit widths on MNIST. Quantile
thresholds are fit over the globally flattened training pixel
distribution, which on MNIST has a point mass at zero exceeding the
$88$\,\% quantile. Inspecting the fitted checkpoint: at $b{=}4$ all
four thresholds are $0.0$, and at $b{=}8$ seven of the eight are $0.0$
with the last at $\approx 0.706$. The ladder therefore collapses: at
$b{=}4$ every bit is the same $(\text{pixel} > 0)$ comparison and the
encoder is effectively $1$-bit, while at $b{=}8$ it is effectively
$2$-bit. Yosys hashes the identical per-bit expressions and emits one
comparator per pixel. The linear thermometer places its thresholds
uniformly in the value range and does not collapse on MNIST, so the
two rows compare a genuine $b$-bit code against an effectively
$1/2$-bit code. That the collapsed variant still leads on accuracy
simply confirms that $(\text{pixel} > 0)$ already carries most of the
MNIST signal.

\section{Hardware evaluation methodology (GPU, FPGA, ASIC)}
\label{sec:appendix_hw_eval}

This appendix documents how every hardware number in
\Cref{sec:hardware_evaluation} is produced: which static-analysis pass
populates each column, how the \ac{GPU}, \ac{FPGA}, and \ac{ASIC} paths
are made comparable, and how the per-mode cycle budget combines with
Vivado's achieved period to yield the reported latency and throughput.
A short closing subsection covers the width-scaling extrapolation that
licences the single-width hardware report.

No physical \ac{FPGA} board is programmed for any reported row: the
\ac{FPGA} numbers come from post-place-and-route Vivado reports or,
where Vivado's placer bails on wide-port designs, from
post-optimization estimates (made precise below). Training runs on a
shared heterogeneous \ac{GPU} cluster (RTX 3090, RTX 2080 Ti, or
Titan RTX) under PyTorch 2.x on \ac{CUDA} 12. The exact card, driver,
and host CPU per seed are recorded in the run record and unioned into
the \textit{GPU host(s)} column of \Cref{tab:training_cost_details} at
aggregation time.

\subsection{GPU methodology}
\label{sec:appendix_hw_gpu}

\ac{GPU} latency, throughput, and energy are measured against a
bit-packed inference path (64 samples per 64-bit operation) that is
graph-captured into a single \ac{CUDA} replay. The remainder of this
subsection describes the path's construction, the compiled forward,
the timing protocol, the telemetry sampler, and the correctness
anchor that ties \ac{GPU} accuracy to the Python evaluation.

\paragraph{Bit-packed inference.}
The \ac{GPU} rows of \Cref{tab:hardware_evaluation} and every accuracy
number in \Cref{sec:sweep,sec:benchmark} use a bit-packed inference
path, not the eager forward. At construction the path walks each
layer once, extracts the discrete $2^n$-entry truth table plus the
per-node routing indices, and normalizes them to a common \ac{LUT}
input bit order. At inference, $64$ samples along the batch dimension
are packed into a single 64-bit integer, and each layer's \ac{LUT}
evaluation becomes a bitwise op on packed tensors that processes $64$
samples per op. The encoder and head remain in floating point. The
packed output is bit-exact with the eager discrete forward and, by
construction, with the emitted SystemVerilog
(\Cref{sec:appendix_hw_hdl_fidelity}). On a modern \ac{GPU} it is
roughly one to two orders of magnitude faster than the eager path.

\paragraph{Compiled forward.}
\label{sec:appendix_hw_processor}
The bit-packed module is wrapped in PyTorch's inference mode and
compiled with graph capture, fusing the per-layer pack, bitwise-\ac{LUT},
and unpack kernels into a single replayable graph. This is the
closest \ac{GPU} analogue of the emitted SystemVerilog: one submitted
unit of work per sample batch, no per-iteration driver round-trips.
The batch is moved to the device once before timing, so no
host-to-device transfers and no dataloader work enter the timed
region.

\paragraph{Timing protocol.}
Per-iteration latency is measured with \ac{CUDA} events placed as
barriers between back-to-back forwards. Warmup fires the forward
without host synchronization until both an iteration-count floor
($500$ iters) and a wall-clock floor ($5$\,s) are met, long enough
for the card's \ac{DVFS} controller to settle out of its idle P-state,
which matters on Turing consumer silicon. A single device
synchronization then gates entry to the timed region, where the driver
records $N{+}1$ barrier events and submits $N{=}5{,}000$ back-to-back
forwards. Per-iteration latency is recovered from the event-to-event
elapsed time. Throughput is $N\cdot\text{batch}$ divided by a
wall-clock bracket around the loop. Each device is benchmarked at
two batch sizes ($64$ and $1024$) so small-batch and sustained
regimes are both visible.

\paragraph{Energy and clock telemetry.}
A background \ac{NVML} sampler polls power, SM clock, memory clock,
and junction temperature at $10$\,Hz over the timed region. Power
samples are integrated to joules with the trapezoidal rule. Clocks
and temperature are summarized as min/mean/max. The card's nominal
boost ceiling is read once at startup, and a row whose observed peak
SM clock sits more than $10\%$ below that ceiling is flagged as
\ac{DVFS}-limited rather than compute-bound, so an artificially
depressed clock is not silently read as a latency win.

\paragraph{Correctness anchor.}
Before timing, the driver compares a one-batch eager forward against
the bit-packed forward: exact equality for LightLUT checkpoints, a
$10^{-4}$ element-wise tolerance for softer parameterizations. Test
accuracy through the bit-packed path is recorded alongside the timing
numbers, so every \ac{GPU} row carries its own correctness check and
is bit-exact with the Python accuracy reported by
\Cref{sec:sweep,sec:benchmark}.

\begin{table}[htb]
\centering
\experitable
\caption{Peak GPU memory (weights plus activations) for the
MNIST winner at two batch sizes. Model-dependent, not card-dependent:
the per-card values coincide, so one multirow lists all three
benchmarked cards.}
\label{tab:hardware_resources_gpu}
\begin{tabular}{@{} l l r @{}}
\toprule
\textbf{Platform} & \textbf{Mode} & \textbf{Memory (MB)} \\
\midrule
\multirow{2}{*}{RTX 3090 / RTX 2080 Ti / TITAN RTX} & batchsize 64 & $7.27$ \\
 & batchsize 1024 & $10.15$ \\
\bottomrule
\end{tabular}

\end{table}

\subsection{FPGA methodology}
\label{sec:appendix_hw_fpga}

The HDL emitter writes one synthesizable SystemVerilog top module
per emission mode and records its cycle budget $(D, \mathrm{II}, C)$,
i.e.\ pipeline depth, initiation interval, and cycles-per-sample. Vivado
then turns those constants and the achieved clock period into the
latency, throughput, and resource numbers reported in
\Cref{tab:hardware_evaluation,tab:hardware_resources_fpga}.

\subsubsection{HDL export and emission modes}
\label{sec:appendix_hw_emit}

Every \ac{FPGA} number originates from a single trained checkpoint.
The HDL emitter extracts each layer's truth table and routing indices
and writes synthesizable SystemVerilog. The top module inlines the
full forward path (encoder comparators, logic layers, and head) as a
single block, with quantized encoder code as input and predicted class
index as output, so every downstream resource number covers the whole
model end-to-end.

Three orthogonal emitter options select the three modes of
\Cref{sec:hardware_evaluation}: encoder style (parallel vs.\
iterative), head style (parallel vs.\ iterative), and three
independent pipeline flags (encoder / logic layers / head). Each mode
fixes a deterministic cycle budget $(D, \mathrm{II}, C)$ that the
emitter records before any synthesis. For an $L$-layer stack, the
core pipeline depth is
\[
D_\text{core} = \mathbb{1}_{\text{pipe\_enc}} + L\cdot\mathbb{1}_{\text{pipe\_layers}} + \mathbb{1}_{\text{pipe\_head}}.
\]
\begin{itemize}
\itemsep0em
\item \textbf{Max throughput.} Parallel encoder and head, with
pipeline flip-flops at the encoder, every logic layer, and the head.
$D_\text{core}{=}4$ on the two-layer MNIST winner, with
$\mathrm{II}{=}1$ (one sample in, one sample out per cycle).
\item \textbf{Lowest latency.} Parallel encoder and head, pipelining
off, so the core datapath is purely combinational ($D_\text{core}{=}0$).
The shim wraps it as a single flop-to-flop stage for Vivado.
\item \textbf{Fewest resources.} Parallel encoder, combinational
logic layers, iterative popcount head. A single shared
popcount/argmax updater walks the $c$ output classes sequentially, so
$\mathrm{II}_\text{core}{=}c$ ($c{=}10$ on MNIST). Iterating the
encoder or layers is a net regression on random-wiring \ac{LUT}
networks at MNIST scale (\Cref{sec:appendix_cost_model}).
\end{itemize}
The full-module rows additionally enable the \ac{BRAM}-backed I/O shim
(\Cref{par:appendix_hw_io_shim}), whose output register adds exactly
one flop-to-flop stage. The cycle-budget calculator folds that $+1$
into $D$, $\mathrm{II}$, and $C$ when the shim is enabled, so the
table renderer needs no shim-specific branch.

\subsubsection{HDL fidelity}
\label{sec:appendix_hw_hdl_fidelity}

For every checkpoint used in \Cref{sec:hardware_evaluation} we run an
equivalence check on the full test split: Verilator compiles the
emitted SystemVerilog with an auto-generated byte-streaming harness,
each test sample is piped through the binary, and its one-byte
class-index output is compared against the bit-packed inference
argmax. All rows across all three emission modes on both \ac{FPGA}
targets report agreement $1.0$ on the full test set, certifying that
every \ac{FPGA} accuracy number is bit-exact with the reported Python
evaluation. The same check runs per seed in the
per-axis sweeps on a $1{,}000$-sample subsample for speed.

\subsubsection{Vivado deployment}
\label{sec:appendix_hw_vivado}

The two \ac{FPGA} targets, an Alveo U55C
(part \texttt{xcu55c-fsvh2892-2L-e}) and a Zynq UltraScale+ XCZU7EV
(part \texttt{xczu7ev-ffvc1156-2-e}), share the UltraScale+
6-\ac{LUT} primitive (so resource counts are directly comparable) and
differ in capacity (${\sim}1.3$\,M \acp{LUT} on U55C vs.\
${\sim}230$\,k on the XCZU7EV). Both run the same non-project Vivado
2024.2 flow, synthesizing out-of-context to skip I/O buffer insertion.

\paragraph{BRAM-backed I/O shim.}
\label{par:appendix_hw_io_shim}
The MNIST winner's $784\times 8 = 6{,}272$-bit encoder input does not
fit either part: the count is far above any Xilinx package's pin
budget and trips Vivado's placer IO-rule check before placement even
begins. To make the design synthesizable \emph{and} make the reported
numbers reflect real evaluation rather than constant-folded ghosts,
the emitter wraps the core in a thin registered shim with a small pin
footprint (clock, reset, class-id). A four-entry \ac{BRAM}-inferred
\ac{ROM} holds distinct sample bit patterns. On every cycle the next
sample is buffered into a wide input register driving the core, and
the core's output is captured one cycle later into an output
register. Because real, varying sample data is clocked through the
network each cycle, Vivado sees genuine flop-to-flop datapaths to
analyse. It cannot constant-fold the \acp{LUT} into a fixed output,
and routed timing and power are reported against meaningful switching
activity. The shim is a synthesis harness, not a deployment
interface. A real system would stream samples through the same
registered input port over AXI-Stream or \ac{HBM}, with
sample-transport latency additive to the reported numbers. The
shim's output register adds exactly one flop-to-flop stage on top of
the core's cycle budget. The emitter folds that $+1$ into $D$,
$\mathrm{II}$, and $C$ when the shim is enabled, so the formulas in
\Cref{sec:appendix_hw_number_recipe} stay shim-agnostic. Only the
full-module row uses the shim. The three submodule runs (encoder,
logic-layer stack, head) stop after optimization, since
\Cref{tab:hardware_resources_fpga} only consumes their
\ac{LUT}/\ac{FF} counts.

\paragraph{Clock constraint.}
Every run that enters timing analysis is constrained against a real
clock (the shim always exposes a clock port), with mode-specific
target periods picked so Vivado meets timing post-route:
\begin{center}
\footnotesize
\begin{tabular}{@{} l r p{0.62\linewidth} @{}}
  \textbf{Mode} & \textbf{$T_\text{clk}$ (ns)} & \textbf{Rationale} \\
  \midrule
  Max throughput     & 8  & Pipelined, $\mathrm{II}{=}1$; achieved critical path $\approx$$7.4$--$7.7$\,ns on both parts, so $8$\,ns leaves $4$--$8\%$ margin. \\
  Lowest latency     & 12 & Single combinational pass through the full forward. \\
  Fewest resources   & 10 & Combinational datapath with an iterative head \ac{FSM}; per-cycle path is short. \\
\end{tabular}
\end{center}
\vspace{-0.5em}
The actual $T_\text{clk}$ is recorded per run so the achieved period
is reconstructed as $T = T_\text{clk} - \text{WNS}$.

\paragraph{Power.}
Power is produced by Vivado's vectorless activity estimation on the
full-module routed netlist (submodule runs skip the power report).
A switching-activity-driven flow would tighten Vivado's
``Medium''-confidence default and is left to future work.

\begin{table}[htb]
\centering
\experitable
\caption{Vivado post-route (\textbf{model}) and post-opt
(\textbf{encoder}, \textbf{layers}, \textbf{head}) \ac{LUT} and
flip-flop utilization per emission mode on the two \ac{FPGA}
targets. The \textbf{model} column reports the whole synthesized
design and is \emph{not} the arithmetic sum of the submodule columns,
because Vivado's synthesizer shares logic across module boundaries.}
\label{tab:hardware_resources_fpga}
\adjustbox{max width=\textwidth}{%
\begin{tabular}{@{} l l r r r r r r r r @{}}
\toprule
 & & \multicolumn{2}{c}{\textbf{model}} & \multicolumn{2}{c}{\textbf{encoder}} & \multicolumn{2}{c}{\textbf{layers}} & \multicolumn{2}{c}{\textbf{head}} \\
\cmidrule(lr){3-4} \cmidrule(lr){5-6} \cmidrule(lr){7-8} \cmidrule(lr){9-10}
\textbf{Platform} & \textbf{Mode} & \textbf{LUT} & \textbf{FF} & \textbf{LUT} & \textbf{FF} & \textbf{LUT} & \textbf{FF} & \textbf{LUT} & \textbf{FF} \\
\midrule
\multirow{3}{*}{U55C} & max throughput & $12{,}515$ & $7{,}857$ & $1{,}568$ & $784$ & $7{,}157$ & $7{,}844$ & $5{,}169$ & $4$ \\
 & lowest latency & $11{,}977$ & $6$ & $1{,}568$ & $0$ & $7{,}461$ & $0$ & $5{,}170$ & $0$ \\
 & fewest resources & $8{,}123$ & $38$ & $1{,}568$ & $0$ & $7{,}461$ & $0$ & $2{,}121$ & $40$ \\
\midrule
\multirow{3}{*}{ZU7EV} & max throughput & $12{,}478$ & $7{,}864$ & $1{,}568$ & $784$ & $7{,}157$ & $7{,}844$ & $5{,}169$ & $4$ \\
 & lowest latency & $11{,}818$ & $6$ & $1{,}568$ & $0$ & $7{,}461$ & $0$ & $5{,}170$ & $0$ \\
 & fewest resources & $7{,}870$ & $38$ & $1{,}568$ & $0$ & $7{,}461$ & $0$ & $2{,}121$ & $40$ \\
\bottomrule
\end{tabular}
}
\end{table}

\subsection{ASIC methodology}
\label{sec:appendix_hw_asic}

The target-independent NAND2-equivalent gate count in
\Cref{tab:hardware_resources_asic} comes from a
Yosys-driven flow against the Nangate 45\,nm open-cell library: read
SystemVerilog, flatten, technology-map against the Liberty file, and
sum post-map cell counts weighted by cell area over the NAND2
reference cell area. The resulting NAND2-equivalent count is a
technology-neutral proxy that lets a reader cross-check against
published numbers from other logic-based \ac{DNN} frameworks. We
deliberately do not attach a gate count to the per-axis sweep tables
of \Cref{sec:sweep}: every axis is swept at three widths, and an
NAND2-equivalent column would be dominated by the linear width
scaling and bury the axis signal.

\begin{table}[htb]
\centering
\experitable
\caption{Target-independent NAND2-equivalent gate count from Yosys
$+$ Nangate\,45\,nm standard-cell synthesis, per emission mode, over
the four columns \textbf{model}, \textbf{encoder}, \textbf{layers},
\textbf{head}. The \textbf{model} column reports the whole
synthesized design and is \emph{not} the arithmetic sum of the
submodule columns, because the area mapper re-balances logic cones
across module boundaries.}
\label{tab:hardware_resources_asic}
\begin{tabular}{@{} l l r r r r @{}}
\toprule
\textbf{Platform} & \textbf{Mode} & \textbf{model} & \textbf{encoder} & \textbf{layers} & \textbf{head} \\
\midrule
\multirow{3}{*}{Nangate 45\,nm} & max throughput & $109{,}073$ & $7{,}840$ & $72{,}623$ & $29{,}471$ \\
 & lowest latency & $63{,}616$ & $3{,}397$ & $28{,}154$ & $29{,}549$ \\
 & fewest resources & $40{,}004$ & $3{,}397$ & $28{,}154$ & $7{,}624$ \\
\bottomrule
\end{tabular}

\end{table}

\subsection{How each number is obtained}
\label{sec:appendix_hw_number_recipe}

Each numeric cell of
\Cref{tab:hardware_evaluation,tab:hardware_resources_gpu,tab:hardware_resources_fpga,tab:hardware_resources_asic}
is the product of a single static-analysis pass and a deterministic
arithmetic derivation. Writing $T = T_\text{clk} - \text{WNS}$ for
the achieved period and
$(D, \mathrm{II}, C)$ for the emitter's cycle budget (on shim-enabled
rows these include the $+1$ flop-to-flop stage from the shim's output
register):
\begin{itemize}
\itemsep0em
\item \textbf{$F_\text{max}$ (MHz).} $F_\text{max} = 1000 / T$.
Identical across all three modes. WNS is the post-route signed slack
against the mode's constraint.
\item \textbf{Latency (ns).} Mode-dependent. With the shim every
mode reduces to a flop-to-flop chain of length $C$, so latency is
uniformly $C\cdot T$: $T$ for lowest-latency ($C{=}1$); $D\cdot T$
for max-throughput ($C{=}D$); $C\cdot T$ for fewest-resources (each
sample occupies the shared \ac{FSM} for $C$ cycles including the
output register).
\item \textbf{Throughput (kSmp/s).}
$\text{Tput} = 10^{3}\cdot F_\text{max} / \mathrm{II}$, with
$\mathrm{II}{=}1$ for lowest-latency and max-throughput and
$\mathrm{II}{=}C$ for fewest-resources.
\item \textbf{Power (W).} Full-module Vivado vectorless estimation on
the routed netlist. Submodule runs do not produce a power cell.
\item \textbf{Energy/smp (nJ).}
$E = 10^{9}\cdot P / (10^{3}\cdot \text{Tput})$, i.e.\ watts divided
by samples-per-second, expressed in nJ.
\item \textbf{LUT / FF.} Vivado utilization report, emitted after
optimization and overwritten by the post-route counts when
place-and-route completes. \ac{DSP} and \ac{BRAM} are zero for the
MNIST winner by construction (fabric-only) and are omitted from
\Cref{tab:hardware_resources_fpga}.
\item \textbf{NAND2-eq.} Yosys + Nangate\,45\,nm flow above,
target-independent, counted over the flattened top module so encoder
comparators and the head output stage are included alongside the
logic layers.
\end{itemize}
The \ac{GPU} rows of \Cref{tab:hardware_evaluation} use the same
single-sample latency (wall-clock through the bit-packed path) and
throughput definitions as the \ac{FPGA} rows, so every row in the
table is directly comparable.

\subsection{Width scaling of hardware metrics}
\label{sec:appendix_hw_width}

Resource and power scale linearly in layer width $w$ with a
width-independent offset, while timing is width-invariant to first
order. The single-width hardware report therefore extends in closed
form. The deployment case study reports a single layer width
($w{=}4{,}000$), the largest cell of \Cref{tab:best_of_space} that
the Vivado synthesis flow admits within our build-host memory budget
on the emitted flat SystemVerilog. The numbers extrapolate to wider
cells of the same two-layer MNIST architecture in closed form: each
logic-layer node emits exactly one $2^n$-entry \ac{LUT} in the
SystemVerilog (one \texttt{localparam} truth table plus one indexed
\texttt{assign} per output neuron), so resource and power counts are
linear in $w$ with a width-independent offset. Writing $L$ for the
number of logic layers ($L{=}2$ on MNIST), $c$ for the number of
classes, and holding fan-in $n$, depth $L$, and emission mode
constant:
\begin{align*}
\mathrm{LUT}(w) &\approx \alpha_\mathrm{LUT}\cdot L\cdot w + \beta_\mathrm{LUT}, &
\mathrm{NAND2}(w) &\approx \alpha_\mathrm{N}\cdot L\cdot w + \beta_\mathrm{N}, \\
\mathrm{FF}(w) &\approx \alpha_\mathrm{FF}\cdot L\cdot w + \beta_\mathrm{FF}, &
P(w) &\approx \alpha_P\cdot L\cdot w + \beta_P,
\end{align*}
where the $\alpha$ coefficients capture per-node contributions (one
\ac{LUT} primitive, one pipeline flip-flop on the inter-layer
boundary in max-throughput mode, the dynamic-power cost of its
switching activity) and the $\beta$ terms collect the fixed-cost
encoder comparator chain, head popcount/argmax tree, and device
static-power floor. The timing path, in contrast, is set by layer
depth $L$ and the head's $O(\log c)$ popcount reduction, \emph{not}
by $w$, because each \ac{LUT}'s $n$ inputs are evaluated independently
and its gate delay does not grow with the number of sibling \acp{LUT}
in the same layer. Accordingly $F_\text{max}$, single-sample latency
$C\cdot T$, and throughput $F_\text{max}/\mathrm{II}$ are
width-invariant to first order, so long as the design fits within the
target part. Energy per sample composes linearly in the same way:
\[
E(w) \;=\; \frac{10^{6}\cdot P(w)}{F_\text{max}/\mathrm{II}} \;\approx\; \frac{10^{6}\cdot\mathrm{II}\cdot(\alpha_P\cdot L\cdot w + \beta_P)}{F_\text{max}}\quad[\mathrm{nJ/sample}].
\]
Empirical support: the rank-2 node sweep underlying
\Cref{fig:node_rank2_nand} reports NAND2-equivalent gate count at
$w\in\{0.5,1,2,4,8,16\}$\,K for LightLUT and recovers a near-linear
fit. The MNIST-winner resource numbers of
\Cref{tab:hardware_resources_fpga,tab:hardware_resources_asic} should
therefore be read as one point on this line. The timing numbers of
\Cref{tab:hardware_evaluation} carry over unchanged for any $w$ that
the target part can host.

\subsection{External reference rows: sources and caveats}
\label{sec:appendix_hw_external}

\paragraph{Sources.}
The two FINN-R MLP-4 rows ($W^1A^1$, MNIST, $97.69\%$ top-1) take
clock and power from Table~5 of \citet{blott2018finnr}. Samples per
second is the reported GOp/s divided by $6.0$\,M ops per frame
(their Table~4), and energy per sample is power divided by samples
per second. The two MLPerf-Tiny IC rows take latency and energy per
inference directly from Table~5 of
\citet{borras2022opensourcefpgamlcodesign}, converted to ns and nJ.
Cells the source paper does not report ($F_\text{max}$ for
MLPerf-Tiny, latency for FINN-R) stay placeholders.

\paragraph{Why they are not directly comparable.}
None of the four rows are a like-for-like comparison with ours. The
FINN-R rows run a binarised \ac{MLP} and the MLPerf-Tiny rows run
CIFAR-10 \acp{CNN}, so the model class differs in every row, and the
dataset also differs in the MLPerf-Tiny rows. Our numbers are Vivado
post-route static estimates, the FINN-R numbers are also predicted
rather than measured, and only the MLPerf-Tiny rows are measured on
real hardware. The reference power figures are board-level and
include the ARM processing system and peripheral rails, while our
U55C and XCZU7EV power numbers are device-level. The rows give an
order-of-magnitude reference, not a head-to-head comparison.

\end{document}